\documentclass{article} 

\usepackage[preprint]{goodfire/goodfire}

\usepackage{mathtools}
\usepackage{amsthm}
\usepackage{multirow}
\usepackage[bottom]{footmisc}

\usepackage{hyperref}
\usepackage{epigraph}
\usepackage{etoolbox}
\usepackage[most]{tcolorbox}
\usepackage{macro}

\usepackage[toc,page]{appendix}
\usepackage[utf8]{inputenc} 
\usepackage[T1]{fontenc}    
\usepackage{hyperref}       
\usepackage{url}            
\usepackage{booktabs}       
\usepackage{amsfonts}       
\usepackage{nicefrac}       
\usepackage{microtype}      
\usepackage[dvipsnames]{xcolor}
\usepackage{bbm}
\usepackage{enumitem}
\usepackage{cancel}
\usepackage{makecell}
\usepackage{tikz}

\usepackage{graphicx} 
\usepackage{subcaption}

\graphicspath{ {./figs/} }
\usepackage{wrapfig}
\usepackage{adjustbox}

\usepackage{amsmath}
\usepackage{amssymb}
\usepackage{todonotes}
\usepackage{natbib}

\usepackage{hyperref}
\usepackage{url}

\usepackage{lineno}

\definecolor{darkblue}{rgb}{0, 0, 0.5}
\hypersetup{colorlinks=true, citecolor=darkblue, linkcolor=darkblue, urlcolor=darkblue}

\usepackage{tcolorbox}
\usepackage{listings}
\providecommand{\aut}[1]{\textbf{#1}}
\providecommand{\af}[1]{{\small #1}}

\providecommand{\afn}[1]{\textcolor{primary}{$^{#1}$}}

\newcommand{\colead}{\textcolor{secondary}{\boldsymbol{\dagger}}}

\newcommand{\goodfireaff}{%
  \includegraphics[height=14pt]{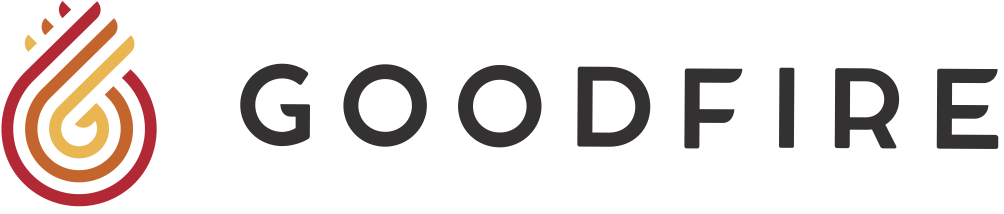}
}

\newcommand{\authorentry}[2]{\aut{#1}\afn{#2}}

\newcommand{\authorsep}{\quad}
\newcommand{\repolink}[1]{%
  {\small
    \href{#1}{\raisebox{-2.8pt}{\includegraphics[height=10pt]{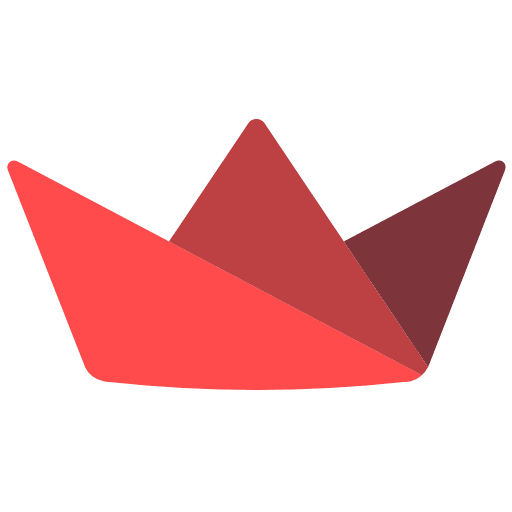}}%
    \texttt{\textcolor{secondary}{#1}}}
  }%
}

\newtoggle{goodfireonly}
\togglefalse{goodfireonly}  

\newcommand{\paperauthors}{%
\authorentry{Eric Bigelow}{} \authorsep
\authorentry{Raphaël Sarfati}{} \authorsep
\authorentry{Daniel Wurgaft}{} \authorsep
\authorentry{Owen Lewis}{} \\
\authorentry{Thomas McGrath}{} \authorsep
\authorentry{Jack Merullo}{} \authorsep
\authorentry{Atticus Geiger}{\colead} \authorsep
\authorentry{Ekdeep Singh Lubana}{\colead} 
\vspace{4pt}\\
\vspace{3mm}
\textcolor{secondary}{$^{\boldsymbol{\dagger}}$}\af{Equal senior contribution} \\
\goodfireaff 
\vspace{3mm}\\
\repolink{~https://conceptual-beliefs.streamlit.app}
\vspace{-4mm}
}
\author{\paperauthors}

\title{Stories in Space: In-Context Learning Trajectories \\ in Conceptual Belief Space}

\begin{document}


\maketitle

\begin{abstract}
    Large Language Models (LLMs) update their behavior in context, which can be viewed as a form of Bayesian inference. However, the structure of the latent hypothesis space over which this inference operates remains unclear.
    In this work, we propose that LLMs assign beliefs over a low-dimensional geometric space---a \textit{conceptual belief space}---and that in-context learning corresponds to a trajectory through this space as beliefs are updated over time.
    Using story understanding as a natural setting for dynamic belief updating, we combine behavioral and representational analyses to study these trajectories. 
    We find that (1) belief updates are well-described as trajectories on low-dimensional, structured manifolds;
    (2) this structure is reflected consistently in both model behavior and internal representations and can be decoded with simple linear probes to predict behavior;
    and (3) interventions on these representations causally steer belief trajectories, with effects that can be predicted from the geometry of the conceptual space.
    Together, our results provide a geometric account of belief dynamics in LLMs, grounding Bayesian interpretations of in-context learning in structured conceptual representations.
\end{abstract}

\section{Introduction}

Large Language Models (LLMs) show impressive abilities to understand language and adapt their behavior according to new information in context~\citep{brown2020language, lampinen2024broader, agarwal2024manyICL, park2024competition, chan2022data, min2022rethinking, park2025iclr, anil2024msj}. 
As an LLM reads a text or engages in a conversation, it must continually update its beliefs about the world, the user, and the current topic being discussed as new information comes to light~\citep{hosseini2026context, geng2025accumulating}.
A pressing question for AI interpretability then is to understand how LLMs represent complex concepts, and how they dynamically update beliefs in these concepts as they process text~\citep{ruis2023llms, hu2025re, prakash2025language, lubana2025priors}.


This dynamic belief updating operates via In-Context Learning (ICL), as LLMs adapt their behavior based on input data, with no changes to model weights. 
%
%
ICL is productively framed as Bayesian inference, where an LLM re-weights latent concepts according to their posterior distributions given input data \citep{xie2021explanation, bigelow2023context, arora2024bayeslaws, panwar2024incontextlearningbayesianprism, zhang2023and}.
Prior work on ICL has primarily focused on few-shot learning scenarios, where data is a list of input--output examples, and where learning dynamics are relatively steady and monotonic \citep{brown2020language, wurgaft2025InContext, bigelow2025belief}.
However, when considering LLM behavior in freeform text generation and understanding, ICL dynamics can be much more non-linear and idiosyncratic~\citep{bigelow2024forking}.
The goal of this work is to extend the Bayesian framework to account for these non-linear ICL dynamics.
In order to accomplish this goal, however, we must consider the structure of latent concepts that an LLM might represent.

\begin{figure}[t]
    \centering
    \vspace{-24pt}
    \begin{subfigure}{\linewidth}
        \includegraphics[width=.9\linewidth]{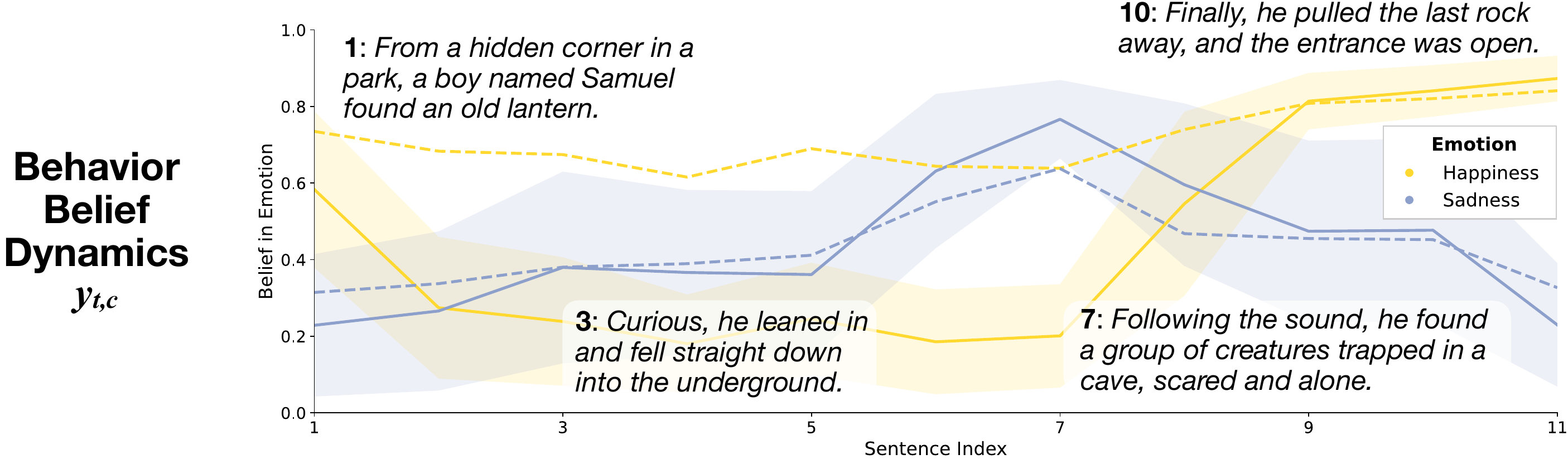}
        \caption{\textbf{Behavior belief dynamics} \ \ As an LLM reads a story $x_{1:T}$, after each sentence $t$ we elicit its beliefs $y_{t,c}$ about a concept $c$ through prompting at that point. 
        These belief dynamics $y_{t,c}$ are shown as a timeseries, with the shaded region showing the standard deviation of $y_{t,c}$ and the dotted lines showing the effect of steering for the concept \textit{happiness}.
        }
        \label{fig:overview-top}
    \end{subfigure}
    %
    \\[10pt]
    \begin{subfigure}{\linewidth}
        \centering
        \includegraphics[width=\linewidth]{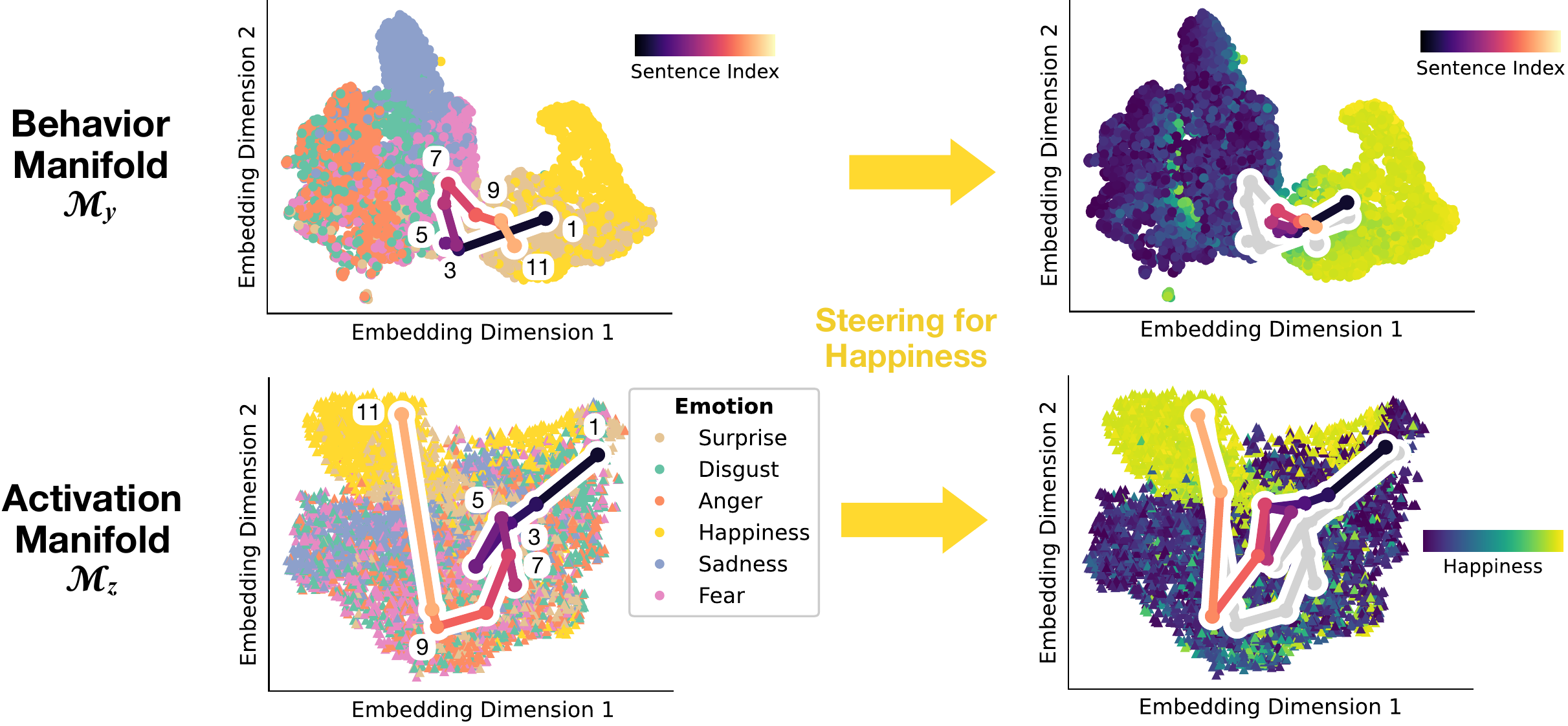}
        \caption{\textbf{Low-dimensional manifolds} \ \ We use dimensionality reduction (UMAP) to identify low-dimensional manifolds in model behavior $\mathcal{M}_y$ and activations $\mathcal{M}_z$.
        These manifolds are structured, with parts shared across behavior and representation, and in-context belief updates following smooth paths through the manifold.
        (Right) The colored path shows the model's trajectory after steering, with the un-steered trajectory shown in grey.
        }
        \label{fig:overview-bottom}
    \end{subfigure}
    \vspace{-5pt}
    \caption{
        \textbf{Conceptual belief trajectories}
        Model beliefs qualitatively follow reasonable patterns for a story, for example, here we see that \textit{happiness} drops when the protagonist falls into a hole and then \textit{sadness} increases when he discovers a group of scared creatures trapped further down, until these reverse at end of the story when the protagonist rescues the creatures (full story text in App.~\ref{app:story-text}). 
        The story's trajectory in (a) behavior space $\mathcal{M}_y$ and (b) activation space $\mathcal{M}_z$ has a similar interpretation, beginning in a positive region, moving towards a negative area, before returning for a happy ending.
        When we steer for a particular concept $c$, in this case \textit{happiness}, we alter the trajectory that the model's beliefs follow.
    }
    \label{fig:overview} 
\end{figure}

%
Since understanding LLMs has much in common with the study of the human mind~\citep{bigelow2026dissertation, hu2025coginterp, hagendorff2023machine}, we draw on cognitive science for theories of conceptual representation in LLMs.
In particular, Bayesian theories of cognition assume that a learner assigns beliefs over latent variables (i.e., \textit{concepts}) which represent a structured generative process that predicts observed data~\citep{tenenbaum2011grow, goodman2008rational, piantadosi2021Computational}.
However, these theories cannot be easily translated to hypotheses about how distributed neural systems implement computations.
To draw connections between patterns in behavior and neural representations, we use the theory of \textit{conceptual spaces} in cogntion, which offers a geometric perspective on concepts as low-dimensional subspaces with well-defined distance metrics~\citep{gardenfors2000conceptual}. 
%
%
Further, we extend this framework to account for notions of uncertainty, or \textit{beliefs}, that a learner might hold over a conceptual space \citep{strossner2022criteria}, and which are essential for Bayesian inference.

In real user interactions, LLMs often process extended, open-ended text where beliefs must be updated as new information is processed.
We use story understanding as a natural setting for studying this kind of belief updating in freeform text, since stories require readers to track events, themes, writer intent, and other information that evolves over time.
Story understanding requires many kinds of knowledge representations and reasoning~\citep{schubert2000episodic, winston2011strong, zwaan1995construction, zwaan1998situation, zacks2001event}, which are dynamically updated in the mind of the reader as they proceed through the story.
We propose that story understanding, and in-context learning more generally, can be viewed as a trajectory through a conceptual belief space.
As a reader acquires knowledge, makes inferences, and predicts each new step of a story, they navigate through a conceptual belief space. 


In this work, we find evidence of exactly such a conceptual belief space in LLMs. We show that (i) LLM behavior and belief updating can be accurately explained by a low-dimensional structured space, and that in-context learning follows smooth trajectories across this space; (ii) we can predict behavior in this space by linearly probing hidden representations; and (iii) we can steer LLMs to change their learning trajectories by intervening on hidden representations.

Overall, our results suggest that, at least in simple cases, LLMs can be shown to track rigorous definitions of belief over time while processing arbitrary freeform text. The Bayesian interpretation of ICL can explain this, although it is more difficult to directly model ICL dynamics since evidence accumulation varies enormously depending on what particular text is used as input data. Instead, we can effectively understand ICL dynamics on freeform text by studying latent structure in conceptual representations which are relevant to the text's content.

\section{In-Context Learning Trajectories in Conceptual Belief Space}

In-context learning is productively viewed as Bayesian inference~\citep{xie2021explanation, wurgaft2025InContext}, where output behavior $y$ is produced by marginalizing over a hypothesis space of latent concepts $c$, each weighted according to how well it matches in-context data $x$:
$$p(y \mid x) = \int_{c} p(y \mid c, x) \ p(c \mid x)$$
However, this framing does not specify what kinds of concepts are represented, and cannot explain a hallmark property of concepts: they are structured in terms of similarity.
As such, we enrich this framing so that the latent concepts belong to a multi-dimensional conceptual space with a metric capturing shared geometry.
Furthermore, we understand ICL belief dynamics as trajectories through a lifted \textit{conceptual belief space} where elements are distributions over the original conceptual space. 

\begin{figure}[t]
    \centering
    \vspace{-12pt}
    \includegraphics[width=\linewidth]{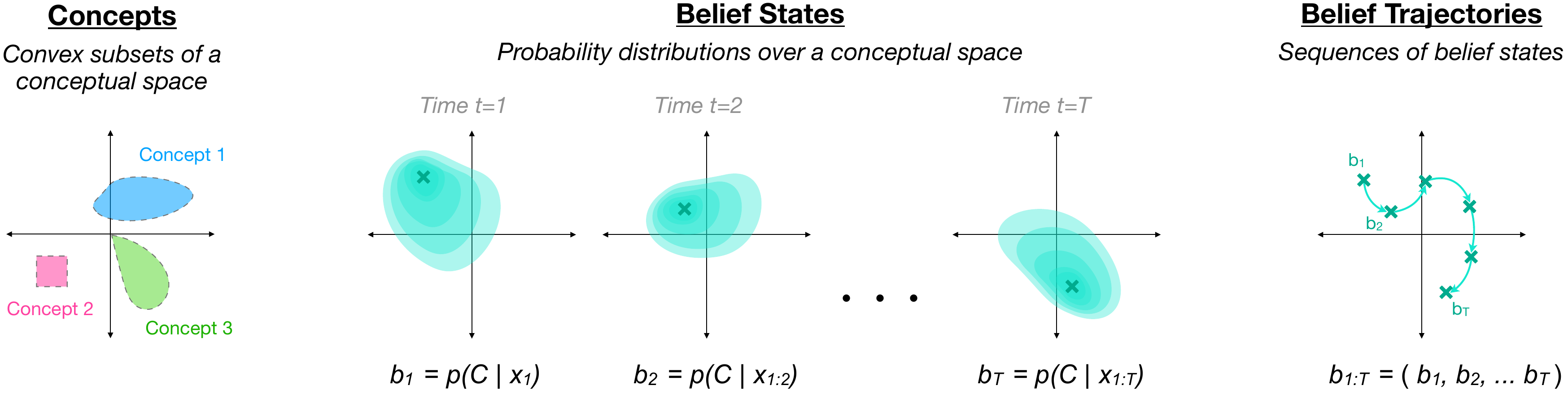}
    \caption{\textbf{Theory of belief update} \quad We extend the conceptual space framework of \citet{gardenfors2000conceptual} to account for belief and uncertainty; axes represent two dimensions in a conceptual space. (Left) Concepts are defined as convex sub-spaces, (Middle) Belief States are probability distributions over a conceptual space, and (Right) Belief Trajectories are sequences of belief states, e.g., in the mind of a reader as they progress through a story~\citep{yeh2025story}.}
    \label{fig:cbs-theory} 
\end{figure}

\paragraph{Conceptual Belief Spaces}
As in \citet{gardenfors2000conceptual}, we define a \textit{conceptual space} as a set of related dimensions and an associated metric, and \textit{concepts} as regions of this low-dimensional geometric space (Fig.~\ref{fig:cbs-theory}). 
%
%
More formally, a conceptual space $\mathcal{C}$ is a metric space over attributes: 
\begin{equation}
    \mathcal{C} = A_1 \times A_2 \times \ldots \times A_n
\end{equation}
Attributes are independent dimensions, on which individual objects and concepts vary.
The structure of this space is defined by a distance metric $d_{\mathcal{C}}$, which measures the similarity of points in $\mathcal{C}$.
%
Attributes in $\mathcal{C}$ can be considered independently of others, for example an object's weight might be determined without considering its temperature and color.
%
%

To move from concepts to beliefs over concepts, we lift to a \textit{conceptual belief space} $\mathcal{B}$, which is a metric space of probability distributions over $\mathcal{C}$~\footnote{See our sister paper \citet{wurgaft2026manifold} for a more rigorous analysis of this framing}:
\begin{equation}
    \mathcal{B} = P(A_1) \times P(A_2) \times \ldots \times P(A_n)
\end{equation}
The dimensions of conceptual belief space are uncertainties over the ground attributes and the metric $d_{\mathcal{B}}$ can be entirely new or a lift of the ground metric $d_{\mathcal{C}}$.

\paragraph{Emotional Belief Space} 
In this paper, we consider a conceptual space over \textit{Emotions}, with the concepts \textit{happiness}, \textit{sadness}, \textit{anger}, \textit{surprise}, \textit{fear}, and \textit{disgust}\footnote{We note the domain we consider for concepts studied in this paper is fairly limited. For example, emotions have been argued to have an intricate, hierarchical organization that results in 135 nodes tree~\citep{shaver1987emotion}}. 
As a reader comprehends each new sentence in a story, such as the story in Fig.~\ref{fig:overview}, they will update their beliefs $p(c \mid x)$ according to whatever new data $x$ they process. For example, when the protagonist endures hardships and conflict, a reader might say that a story seems more \textit{sad}, and then when they return victorious from a great journey, the story might seem \textit{happy}. 
We will behaviorally operationalize this by simply asking a reader (in our case, an LLM) about a particular concept $c$ after each sentence \textit{``How (happy/sad) is this story, on a scale of 0-10?''} of the story and recording their behaviors as  $p(y_{\text{sad}} \mid x)$ and $p(y_{\text{happy}} \mid x)$. 
This process of dynamic change in emotional tone can be visualized as a line plot for each emotion, shown in Fig.~\ref{fig:overview}, which tracks the ups and downs of the reader's beliefs during each twist and turn of the story~\citep{vonnegut1995}.
%
%

We predict that the \textit{Emotions} and \textit{Genres} domains will have structured representations in conceptual belief space $\mathcal{B}$ according to relationships between different concepts that occupy the space.
One leading theory in human psychology proposes that all emotions fall along the two related dimensions of \textit{Valence} and \textit{Arousal}~\citep{russell1977evidence, russell1980circumplex}.
%
%
This theory would lead us to predict that if we asked people to judge various stimuli according to emotions such as \textit{happiness}, \textit{sadness}, or \textit{anger}, then both their behavior and neural representations would follow certain patterns in accordance with their distances $d(c, c')$ in valence-arousal space~\citep{balkenius2016spaces}.

\paragraph{Trajectories through Conceptual Belief Space} We define \textit{belief states} $b_t$ that an LLM might have at a particular point in a story $t$ as a probability distribution over a conceptual space $\mathcal{C}$, i.e., $b_t = p(\mathcal{C} \mid x_{1:t}) = p(A_1 \ldots A_n \ | \ x_{1:t})$ where $x_{1:t}$ refers to a story text $x$ from its beginning up until point $t$. 
These states occupy a wider \textit{conceptual belief space}.
Finally, a \textit{belief trajectory} $b_{1 : T}$ is a path through conceptual belief space, i.e., a sequence of belief states for each time $t$ throughout a story $x_{1:T}$: $b_{1 : T} = \big( b_1, b_2, \ldots b_T \big)$.
%

%
This framework will enable us to talk formally about how a reader incrementally updates their beliefs while processing a story, and about the structure of latent concepts that they use to assign beliefs. 
%
%
%
Concretely, we predict that LLM behavior and representations can be described according to positions in this lower-dimensional conceptual belief space, with minimal loss of information.
%
An advantage of this framework is that psychological spaces may be mapped onto dimensions of the underlying neural substrate, for example, in humans, emotions might map to different quantities of neurotransmitters such as dopamine and seratonin~\citep{wang2020neurotransmitters}, and colors can map to rods and cones in the retina~\citep{brown1964visual}.
%
%
However, in many cases we may not have a theory a priori of what underlying dimensions define a particular conceptual space. Instead, we can use dimensionality reduction techniques to infer the latent dimensions given behavioral or neural data for different concepts.




%


\section{Methods}

We next provide a set of methods for studying belief update dynamics in LLMs and the low-dimensional conceptual spaces which underlie these beliefs.
For the following methods, we will be studying how an LLM updates its beliefs as it reads each successive sentence $x_{t}$ of a text $x_{1:T}$. We will collect two kinds of data: behavioral judgments $p(y_c \mid x_{1:t}, q_c)$ by the LLM which quantify how well a given concept (e.g. \textit{happiness} or \textit{sadness}) applies to $x_{1:t}$, and residual activations for the story text $z_\ell(x_{1:t})$ for layer $\ell$ at the final token of $x_{1:t}$.
%
%
%
For behavioral judgments, we prompt the LLM to rate how well a \textit{concept} applies to $x_{1:t}$. We do not assume that we know a priori the underlying conceptual dimensions $\{A_1, \ldots, A_n\}$. Instead, we assume that we know a set of related concepts (e.g. different \textit{Emotions} or \textit{Genres}) which should fall under some domain $\mathcal{D}$.

\begin{figure}[t]
    \centering
    \includegraphics[width=\linewidth]{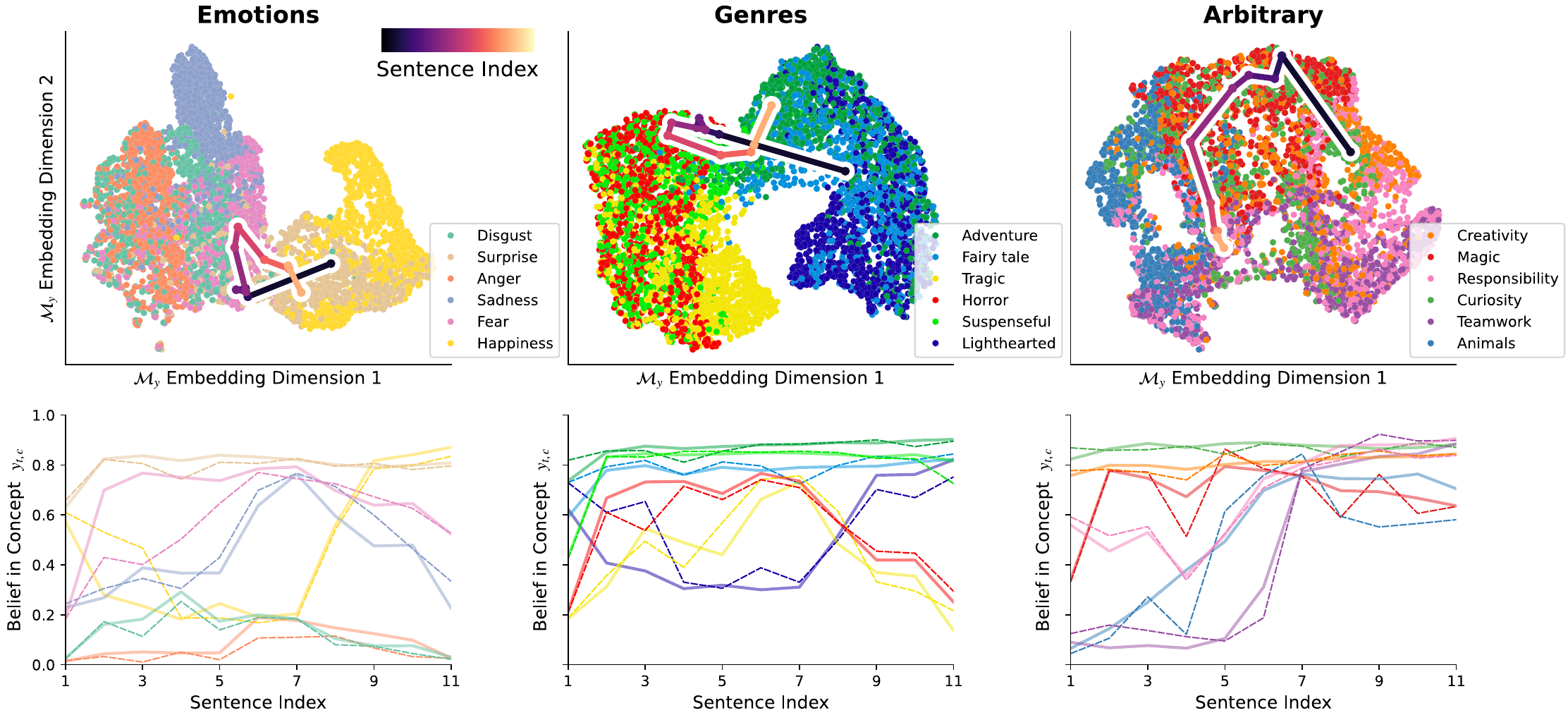}
    \vspace{-20pt}
    \caption{
        \textbf{Three conceptual domains} \ \
        (Top) Belief manifolds $\mathcal{M}_y$ across three domains, with the trajectory plotted for the same story as in Fig.~\ref{fig:overview}.
        For both \textit{Emotions} and \textit{Genres}, the more ``positive'' concepts (in terms of valence) are on one end of the manifold, with negative emotions at the opposite end, and the more neutral concepts (\textit{surprise} and \textit{adventure}) forming a bridge between the two. 
        On the other hand, $\mathcal{M}_y$ for \textit{Arbitrary} does not have such clear structure.
        (Bottom) Observed model beliefs $y_{t, c}$ (solid line) plotted against probe predictions $\widehat{y}_{t, c}$ at layer $\ell=9$.
        We find that linear probes are highly predictive of model behavior, suggesting that belief is encoded linearly as a model processes story text.
    }
    \label{fig:three-domains} 
\end{figure}

\paragraph{Dimensionality Reduction}

We predict that behavioral data $y$ and representational data $z$ can be described according to movement along low-dimensional manifold $\mathcal{M}_y$ and $\mathcal{M}_z$. To approximate these manifolds, we use dimensionality reduction techniques---UMAP~\citep{mcinnes2018umap} and PCA---on our datasets $Y \in \mathbb{R}^{N \times k}$ and $Z_\ell \in \mathbb{R}^{N \times q}$.
Important information should be preserved after projecting the LLM's beliefs into the lower-dimension spaces $\mathcal{M}_y, \mathcal{M}_z $.
%
To characterize the outer edges of the model's belief manifold, we fit dimensionality reduction to a set of the most extreme examples of each concept, and project data points onto this space. 
We filter our training dataset to the sentences across any story which maximally activate each concept, for example, in the \textit{Emotions} domain we take the \textit{``happiest''} points in any story, the \textit{``saddest''} points, and so on. 
%
%
%
We will also empirically test for $\mathcal{M}_y$ and $\mathcal{M}_z$ will be jointly predictive of both model behavior and internal representations, and that structure in effects of activation steering. 
Both $\mathcal{M}_y$ and $\mathcal{M}_z$ can be seen as images of an underlying conceptual belief space $\mathcal{B}$.
%
%
More broadly, since $y_{t,c}$ is an overcomplete representation that lies on or near the manifold $\mathcal{M}_y$, $y_{1:T}$ can be visualized by projecting it into the lower-dimensional space $\mathbb{R}^{T \times d}$.

We make three concrete predictions about $\mathcal{M}$: (1) belief in the  \textit{Emotions} and \textit{Genres} domains can be well-described by coordinates in $\mathcal{M}_y$ and $\mathcal{M}_z$; manifolds for the \textit{Emotions} and \textit{Genres} domains will be structured according to the relationships of different concepts $c$, for example emotions may have hierarchical~\citep{plutchik1980general} or geometric~\citep{russell1980circumplex} structure as has been shown for humans; (3) behavior $\mathcal{M}_y$ and activation $\mathcal{M}_z$ manifolds will be similar in structure.

\paragraph{Distance Matrices}    ~\label{sec:methods-distance}
In order to empirically test these hypotheses, we will use distance matrices (or, equivalently, similarity matrices) with an $L_2$ metric to quantify structure and similarity of $\mathcal{M}_y$ and $\mathcal{M}_z$, somewhat similar to RSA \citep{kriegeskorte2008representational}. 
Since each concept refers to a region in $\mathcal{M}$, we will use the centroid for a concept in the embedding space to approximate it's geometry as a point. More specifically, for each concept $c$ we will compute the mean coordinates over a set of max-activating examples $x_{1:t}$, i.e., $\frac{1}{N} \sum_n \mathcal{M}(x_{1:t}^{(n)})$. In simple terms, for a two-dimensional space $\mathcal{M}$ will be represented merely with a single coordinate pair.
Each point in a distance matrix for a manifold $\mathcal{M}$ will measure the distance $d_\mathcal{M}(c, c')$ between the centroids of two concept $c$ and $c'$.

After constructing distance matrices, we will perform two kinds of analyses to measure the structure and similarity between two geometric spaces.
First, we will use agglomerative clustering~\citep{ward1963hierarchical} to infer hierarchical structures which may be present in distance matrices. 
Second, we will measure correlation between distance matrices, by comparing $d(c, c')$ across different distance functions $d$.


\begin{figure}[t]

    \centering
    \includegraphics[width=\linewidth]{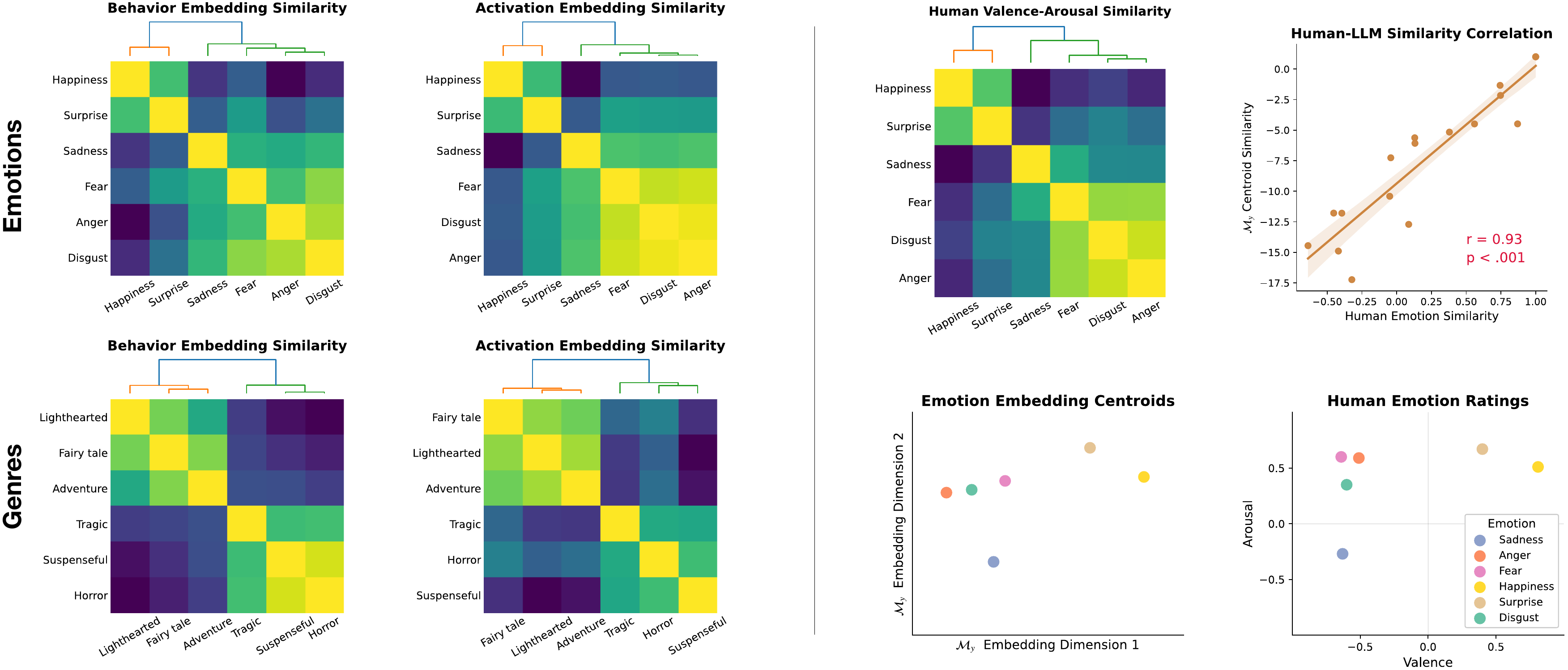}
    \caption{\textbf{Distance reveals domain structure} \ \  (Left) Distance matrices for pairs of concepts $d_\mathcal{M}(c, c')$ according to behavior $\mathcal{M}_y$ and activation $\mathcal{M}_z$ manifolds, computed between concept centroids.
    For \textit{Emotions} and \textit{Genres}, we find common hierarchical structure shared across $\mathcal{M}_y$ and $\mathcal{M}_z$, and that concept centroid distances $d_\mathcal{M}(c, c')$ are highly correlated between $\mathcal{M}_y$ and $\mathcal{M}_z$ ($r=.92, p<.001$ for \textit{Emotions}, $r=.89, p<.001$ for \textit{Genres}). Results for \textit{Arbitrary} shown in App.~\ref{app:arbitrary-analysis}.
    (Right) The geometry of $\mathcal{M}_y$ for the \textit{Emotions} domain is strikingly similar to the geometry of valence-arousal emotion space estimated from human data \citep{russell1977evidence}.
    }
    \label{fig:distance-mats} 
    
\end{figure}

\paragraph{Belief Elicitation} ~\label{sec:beh-elicit}
Behavioral observations offer a lens into how an LLM updates its beliefs dynamically over the course of a story.
To operationalize this, we input a story text into an LLM up until some time $t$, i.e., $x_{1:t}$, and then append a query $q_c$ to the LLM's input text which asks the LLM to rate on a scale from 0 to 10 how much a concept $c$ applies to the text $x_{1:t}$.
We prompt models to respond with an integer in the range 0 to 10, $i \in [0, 1, \ldots 10]$ (full prompts listed in App.~\ref{app:query-prompts}), and we take their estimate $y_{t, c}$ as a weighted average of their token probabilities for each possible integer value, i.e.:
\vspace{-5pt}
\begin{equation}
    y_{t, c} = \mathbb{E}[y = i_c] =
        \frac{1}{10} \sum_{i=0}^{10} \ i \cdot p(y = i \ | \  x_{1:t}, q_c)
\end{equation}
Each $y_{t,c}$ can be considered as a behavioral estimate of the model's belief in concept $c$ at time $t$.
We repeat this procedure for each queried concept $q_c$, and the behavioral belief state $y_t$ is comprised of all queried concepts: $y_t = \{ y_{t, c_1}, y_{t, c_2}, \ldots  y_{t, c_k}\}$. 
By concatenating this sequence of belief probabilites $y_t = p(a \ | \  x_{1:t})$ for each timestep, we form a belief trajectory $y_{1:T} = (y_1, y_2, \ldots y_T)$ across the entire span $T$ of a story. Iterating across a dataset of stories, we collect behavioral data $Y \in \mathbb{R}^{N \times k}$ where $k=6$ in our experiments.

\paragraph{Linear Probes} ~\label{sec:probes-defn}
Next, we use probes~\citep{alain2017understanding, hewitt-liang-2019-designing, conneau2018you, tenney2019bert, belinkov2017neural, belinkov2022probing, tenney2018what} to predict an LLM's output behavior $y_{t, c}$ from its hidden activations $z_{t, \ell}$.
To collect activations, we input the text of a story $x_{1:t}$ from the beginning until sentence $t$, and collect residual activations $z_{\ell, t}$ at each layer $\ell$ for the final token. Only the raw content of the story $x_{1:t}$ is presented as input to the LLM, without any specific concepts mentioned.
Given dataset of activations $Z_\ell \in \mathbb{R}^{N \times q}$ (where $q=4096$ in our case) across stories, we fit linear probes to predict behavioral beliefs $y_{t, c}$: $\widehat{y}_{t, a} = f_{\theta} (z_{t, \ell}) $, using linear regression $f_{\theta}(z) = \theta^\top z + \theta_0$ with $L_2$ sparsity penalty.
Lastly, we fit isotonic regressions to calibrate probes and map $f_{\theta}(v)$ onto the scale $[0, 1]$ (App.~\ref{app:calibration})~\citep{niculescu2005predicting}.
Probes are trained and calibrated on a larger training dataset and tested on a held out test set.
%

%
%
%
%

\paragraph{Activation Steering}

We next aim to verify that the representations we identify are causally involved in determining behavior. To do so, we intervene on latent activations in an LLM to influence its beliefs. The class of methods we will use is activation addition steering, in other words, methods where a steering vector is constructed for a given layer and concept $v_{c, \ell}$. Activations are steered at a specific layer by intervening on the LLM during its forward pass and patching activations $z_{\ell}(x_{1:t})$ to add in the steering vector, multiplied by some scalar magnitude $\alpha$, i.e., $\widetilde{z_{\ell}}(x_{1:t}) = z_{\ell}(x_{1:t}) + \alpha \ v_{c, \ell}$.
We use two primary methods for activation steering: linear probe weights and difference in means steering vectors. For linear probes, we directly use the normalized parameter vector $\frac{\theta}{\| \theta \|}$ for a given attribute. Difference-in-Means (DIM) vectors~\citep{bau2018identifying, panickssery2024Steering,  turner2024Activation, marks2024Geometry, rimsky-etal-2024-steering}, are constructed by taking the mean activations for two contrasting datasets $v_{c, \ell} = \overline{z_\ell (x_c)} - \overline{z_\ell (x_{\neg c})}$.
%


%


\section{Experiments}

We use the SimpleStories dataset \citep{finke2025parameterized}, a collection of synthetically generated short stories with various themes and styles. In our analyses, we use a training dataset of $2500$ randomly sampled stories, with $500$ stories each for each of the following ``style'' categories: \textit{adventure}, \textit{classic}, \textit{lighthearted}, \textit{melancholy}, and \textit{tragic}, as well as a held-out test dataset with $500$ stoies.
For analysis, the conceptual domains we use are \textit{Emotions}, \textit{Genres}, and \textit{Arbitrary}, and for each domain we test between 6 concepts. These domains were chosen such that Emotions and Genres should be expected to have a latent conceptual structure, such as hierarchy, except for the Arbitrary domain which serves as a counter-example and is expected to have less clear structure, and less predictive low-dimensional embeddings.
For dimensionality reduction, we filter our training dataset to the $1000$ sentences which maximally activate each concept, with a maximum 3 data points for a single story. 
The experiments presented here all use \texttt{llama-3.1-8b-instruct} model~\citep{grattafiori2024llama} with 4-bit weight quantization for efficiency~\citep{dettmers2023qlora}.

\begin{figure}[t]
    \centering
    \includegraphics[width=\linewidth]{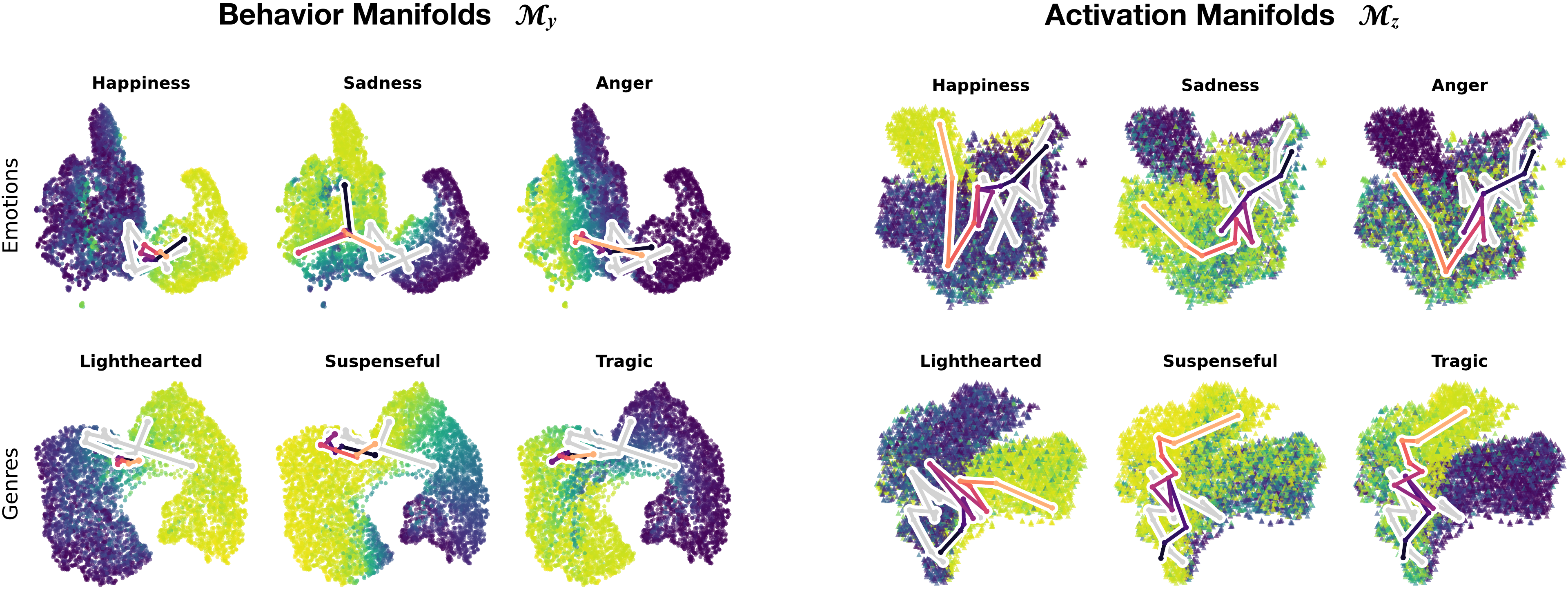}
    \caption{\textbf{Steering along manifolds} \ \ When we steer the story shown in Figs.~\ref{fig:overview}~\ref{fig:three-domains}, the story's trajectory in belief space shifts towards the parts of the manifold that corresponds to a particular concept.
    Each figure here shows the effect of steering the trajectory $b_t$ towards a particular concept $c$, along with the manifold $\mathcal{M}_y$ or $\mathcal{M}_z$ with each point colored by the value $y_{t,c}$ for that concept.
    Results for all concepts listed in App.~\ref{app:steer-everything}.
    }
    \label{fig:steer-geom}
\end{figure}

\section{Results}

\paragraph{Behavioral beliefs can be linearly decoded from activations}
We find rich behavioral dynamics in \texttt{Llama-3.1-8b}~\citep{grattafiori2024llama} for many attributes in nearly all stories in our dataset, which follow the semantic content of each story in subjectively reasonable ways (Figs.~\ref{fig:overview},~\ref{fig:three-domains})~\footnote{An interactive dashboard with belief dynamics and trajectories for all stories in our test set is available at: \url{https://conceptual-beliefs.streamlit.app}}. 
Our probes at layer $\ell=9$ are able to predict behavioral belief dynamics with high accuracy (Figs.~\ref{fig:three-domains}), achieving $.09$ RMSE on the \textit{Emotions} domain, $.09$ RMSE on the \textit{Genres} domain, and $.11$ RMSE on the \textit{Arbitrary} domain (see App.~\ref{app:probe-opt-layer} for analysis across layers).
These results support our claim that the behavioral signatures we find are reflective of beliefs that the LLM holds while reading a story, since we can predict behavior from the model's activations even when it is not instructed to consider specific concepts.

\begin{figure}[t]
  \vspace{-20pt}
  \centering
  \includegraphics[width=\linewidth]{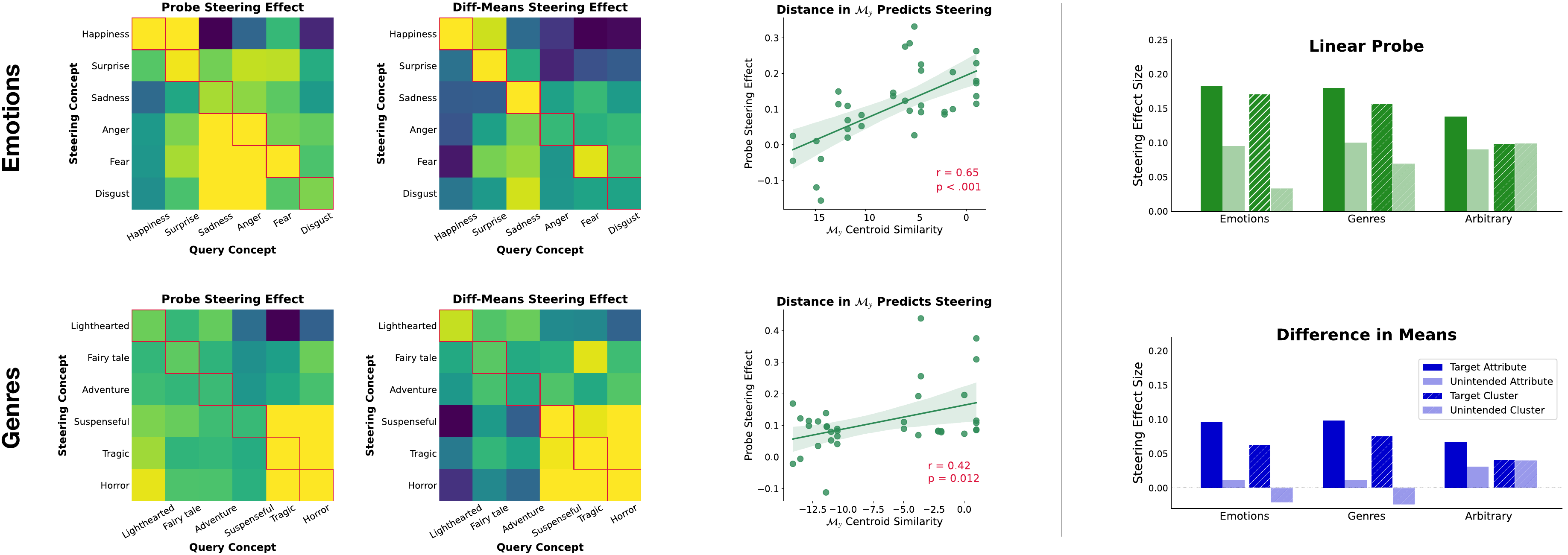}   
  \caption{\textbf{Steering entanglement follows manifold geometry} \ \ 
      We find that steering for a target concept $c$ (like \textit{sadness}) often increases belief in related but unintended concepts $c'$ (like \textit{anger}). 
      \textbf{(Left)} Steering effect, or change in $y_{t, c}$, when steering for a particular target concept $c$ (y-axes on heatmaps), and measuring belief in query concept $c'$. \textit{Emotions} domain is on top, \textit{Genres} on bottom.
      \textbf{(Middle)} Steering effect can be predicted from embedding similarity $d_{\mathcal{M}_y}$.
      %
      %
      \textbf{(Right)} Steering effect is significantly higher for the target concept $c$ compared to the mean effect across other concepts $c'$. However, when we cluster concepts according to the top level of the hierarchies in Fig.~\ref{fig:distance-mats}, the effect on the ``unintended'' cluster disppears for the \textit{Emotions} and \textit{Genres} domains. We see lower on-target effects and higher off-target effects in the manifold of \textit{Arbitrary} concepts, for which the concepts do not share geometric structure.
  }
  \label{fig:steering-entangle}
  \vspace{-10pt}
\end{figure}

%
%
%

\paragraph{Shared structure in low-dimensional belief and representation manifolds}

We find strong qualitative and quantitative structure in the geometry of $\mathcal{M}_y$ and $\mathcal{M}_z$ (Fig.~\ref{fig:three-domains}) for the \textit{Emotions} and \textit{Genres} domains, and belief trajectories $b_{1:T}$ follow relatively smooth and interpretable paths.
%
%
%
%
%
%
Behavioral embeddings $\mathcal{M}_y$ closely follow activation embeddings $\mathcal{M}_z$ (Fig.~\ref{fig:overview}), particularly for the \textit{Emotions} and \textit{Genres} domains. This is true even when we only compare two-dimensional centroids in $\mathcal{M}_y$ for each concept.
In order to measure similarity between low-dimensional spaces, we map data points in these spaces to distance matrices, as shown in Fig.~\ref{fig:distance-mats} (refer to Sec.~\ref{sec:methods-distance} for details).
%
%
We find that the resulting distance matrices are highly correlated between values embedded into the behavior manifold $\mathcal{M}_y$ and activation manifold $\mathcal{M}_z$ (Pearson's $r=.92, p<.001$ for \textit{Emotions}, $r=.89, p<.001$ for \textit{Genres}, and $r=.71, p<.001$ for \textit{Arbitrary}). 
We find that the hierarchies inferred in Fig.~\ref{fig:distance-mats} directly correspond to our qualitative observations of $\mathcal{M}_y$, and inferred hierarchies are equal between $\mathcal{M}_y$ and $\mathcal{M}_z$ for the \textit{Emotions} and \textit{Genres} domains.

Finally, we compare the embedding centroids we learn for the \textit{Emotions} domain to human data from \citet{russell1977evidence}, where people rated emotion words according to Valence and Arousal, and Dominance~\footnote{We limit the space to Valence and Arousal in line with circumplex theory~\citep{russell1980circumplex} and to have a two-dimensional space.} (Fig.~\ref{fig:distance-mats},~Right). We find that our centroid embeddings are a close match for human emotion ratings on the dimensions of Valence and Arousal, and that the distance matrix for the human data is highly correlated with $\mathcal{M}_y$ and $\mathcal{M}_z$ ($r = .93, p<.001$ for $\mathcal{M}_y$), complementing prior work demonstrating similarity of emotion organization in language models and humans \citep{zhao2025emergence, sofroniew2026twheemotion, sun2026valence}.

\paragraph{Manifold geometry predicts steering effects}

We find that activation steering with linear probes and difference-in-means are effective in steering LLM beliefs (Fig.~\ref{fig:steering-entangle}). With linear probes, we found that steering with only a single layer was ineffective, since later layers would correct for the intervention and nullify its effects (App.~\ref{app:probe-effect-repair}). Steering simultaneously with vectors across a span of $7$ layers corrected for this.
With difference-in-means steering we did not encounter this issue, and instead found steering at a single layer to be effective. 

Next, we found that steering for one concept often leads to a change in behavior for other concepts (Fig.~\ref{fig:steering-entangle}).
Crucially, steering entanglement between concepts mirrors the geometry of the conceptual domains, e.g. steering for a negative emotion such as \textit{sadness} leads to an increase in \textit{anger} ratings and a decrease in \textit{happines} ratings.
We estimated steering entanglement by comparing the effect on the targeted concept to the average effect of steering on our other measured concepts.
Comparing across steering methods, we find that difference in means steering, compared to linear probes, have less effect on unintended concepts, but they also have a smaller maximum effect on the target concept (Fig.~\ref{fig:steering-entangle}, Right).

We also find that steering entanglement can be predicted based on the structure of our learned manifold $\mathcal{M}_y$. Steering effect between two concepts is correlated with the distance between their centroids in $\mathcal{M}_y$ ($r=.65, p<.001$ for \textit{Emotions}, $r=.42, p=.01$ for \textit{Theme}; Fig.~\ref{fig:steering-entangle}, Center Left).
We can also use a linear model to predict out-of-sample the degree of effect that steering for one concept will have for another queried concept (i.e., $c$ in $q_c$; Fig.~\ref{fig:steering-entangle}, Center Right).
Finally, we re-analyzed steering effect on target or unintended concepts by using the top-level clustering of the dendrograms that we infer for $\mathcal{M}_y$ to measure the steering effect of concepts within a cluster, against the steering effect on concepts outside this cluster. E.g. in the emotion domain, we have two clusters, one with $\{ \textit{happiness}, \textit{surprise} \}$ and one with the other four emotions.


\section{Discussion}

%
Our results broadly support our theory that LLMs operate over low-dimensional \textit{conceptual belief spaces} which group into structured domains of related dimensions.
We find support that our behavioral probe is qualitatively working as expected by examining $y_{t, c}$, and that this can be predicted from $z_t$ with linear probes suggests that the same information is represented in the LLM's residual stream when it only processes the story text $x_{1:t}$.
The manifolds we learn $\mathcal{M}_y$, $\mathcal{M}_z$ have rich structure, and where behavior $\mathcal{M}_y$ and activation $\mathcal{M}_z$ manifolds are highly correlated and share the same hierarchical structure for the domains of \textit{Genres} and \textit{Emotions}~\citep{sofroniew2026twheemotion, zhao2025emergence, dong2025controllable}.
We can use linear activation steering to alter model behavior and representations, shifting the in-context learning trajectories for different stories along $\mathcal{M}_y$ and $\mathcal{M}_z$. 
We also observe steering entanglement across concepts, where steering for one concept unintentionally also shifts model beliefs for another concept, and this steering entanglement can be predicted based on the hierarchical structure in $\mathcal{M}_y$.

%
Our work is a first step towards understanding conceptual belief spaces in LLMs, and for developing a principaled theoretical framework which unifies geometric theories of conceptual representation with probabilistic theories of learning.
We hand-select three domains with six concepts each, but our method is limited by how we chose domains and concepts based on prior knowledge. This parallels the hand-engineering of a concept space in Bayesian cognitive science~\citep{tenenbaum2011grow}, although like with these methods~\citep{wong2023Word}, a promising direction for future work will be to apply ``auto-interpretability'' methods~\citep{bills2023language, paulo2024automatically} for proposing conceptual domains for a given LLM task.
Another limitation is that the concepts we test are treated as unitary and non-compositional, and are also global for the story as a whole, rather than applying to e.g. a particular character in a story. We hope to extend these methods in future work with much richer conceptual belief spaces, with more elaborate concepts.
Finally, another question we hope to explore in future work is how conceptual belief spaces shape uncertainty in long-form text generation~\citep{bigelow2024forking, roadnottaken, ahdritz2024distinguishing, boppana2026reasoning, stolfo2024confidence, han2024semantic}, in addition to language understanding.

%
This work represents an early step towards a theoretical foundation for LLM interpretability which unifies principles of Bayesian program learning \citep{tenenbaum2011grow, lake2017building, ullman2020bayesian} and conceptual spaces \citep{gardenfors2000conceptual, gardenfors2001reasoning} in cognitive science. However, much remains to be done in further solidifying this foundation.
For example, our results raise the questions of what effect different kinds of activation steering could have on belief trajectories, and whether a new steering could be developed which takes advantage of the manifolds $\mathcal{M}$.
%
%
Our framework also raises the question of how, specifically, these conceptual spaces relate to sophisticated behaviors involving likelihood $p(c \mid x)$ and predictive distributions $p(y \mid c)$.

\newpage
\bibliography{refs}
\bibliographystyle{plainnat}

\newpage
\appendix

\section{Story Text}  ~\label{app:story-text}

Here in Fig.~\ref{fig:story-text}, we provide the full text for the story shown in Figs.~\ref{fig:overview},~\ref{fig:three-domains},~\ref{fig:steer-geom}.

\begin{figure}[h!]
    \centering
    \begin{tabular}{ c|l } 
     $i$  & \hspace{100pt} \textbf{Sentence} \\ \hline \hline
     1 & \small{\textit{From a hidden corner in a park, a boy named Samuel found an old lantern.}}  \\ \hline
    2 & \small{\textit{When he turned it on, a bright light shot out, revealing a dark hole in the ground.}}  \\ \hline
    3 & \small{\textit{Curious, he leaned in and fell straight down into the underground.}}  \\ \hline
    4 & \small{\textit{The air was thick and smelled of damp earth.}}  \\ \hline
    5 & \small{\textit{As he stood up, he saw he was in a strange world where time seemed to stop.}}  \\ \hline
    6 & \small{\textit{But soon, he heard cries for help.}}  \\ \hline
    7 & \small{\textit{Following the sound, he found a group of creatures trapped in a cave, scared and alone.}}  \\ \hline
    8 & \small{\textit{Thinking quickly, Samuel gathered rocks and started to clear the entrance.}}  \\ \hline
    9 & \small{\textit{The creatures cheered as he worked.}}  \\ \hline
    10 & \small{\textit{Finally, he pulled the last rock away, and the entrance was open.}}  \\ \hline
    11 & \small{\textit{\makecell[l]{The creatures ran out, thanking him for his bravery, while Samuel \\ realized he had found true friends in an unexpected place.}}}  \\ \hline
    \end{tabular}
    \caption{Full text for story in Figs.~\ref{fig:overview},~\ref{fig:three-domains},~\ref{fig:steer-geom}, split into numbered sentences.}
    \label{fig:story-text} 
\end{figure}

\newpage
\section{Belief Query Prompt}  ~\label{app:query-prompts}

In the following, we present the specific prompts $q_c$ that we used to behaviorally query for the model's beliefs $y_{t,c}$ in specific concepts.

\begin{tcolorbox}[colframe=teal, title=Emotion Query Prompt]
\begin{lstlisting}
Consider the following story:
```
{story}
```

Now, your task is to determine the emotional content of this story, specifically for the emotion {emotion_noun}.
Respond with only a single word, an integer in the range [0, 10], where 0 is not at all {emotion_adjective}, 5 is neutral, and 10 is the most {emotion_adjective}.
What is the level of {emotion_noun} of this story?
\end{lstlisting}
\end{tcolorbox}

\begin{tcolorbox}[colframe=teal, title=Genre Query Prompt]
\begin{lstlisting}
Consider the following story:
```
{story}
```

Now, your task is to determine the thematic content of this story, specifically for the theme "{genre}".
Respond with only a single word, an integer in the range [0, 10], where 0 is not at all {genre}, 5 is neutral, and 10 is the most {genre}.
How much does this story match the theme "{genre}"?
\end{lstlisting}
\end{tcolorbox}

\begin{tcolorbox}[colframe=teal, title=Arbitrary Query Prompt]
\begin{lstlisting}
Consider the following story:
```
{story}
```

Now, your task is to determine whether the concept "{concept}" applies to this story.
Respond with only a single word, an integer in the range [0, 10], where 0 means the concept is very unlikely, 5 is neutral, and 10 means the concept is very likely.
Does the concept "{concept}" apply to this story?
\end{lstlisting}
\end{tcolorbox}

\newpage
\section{Steering Effects for All Concepts and Domains}  ~\label{app:steer-everything}

\begin{figure}[t]
    \centering
    \includegraphics[width=\linewidth]{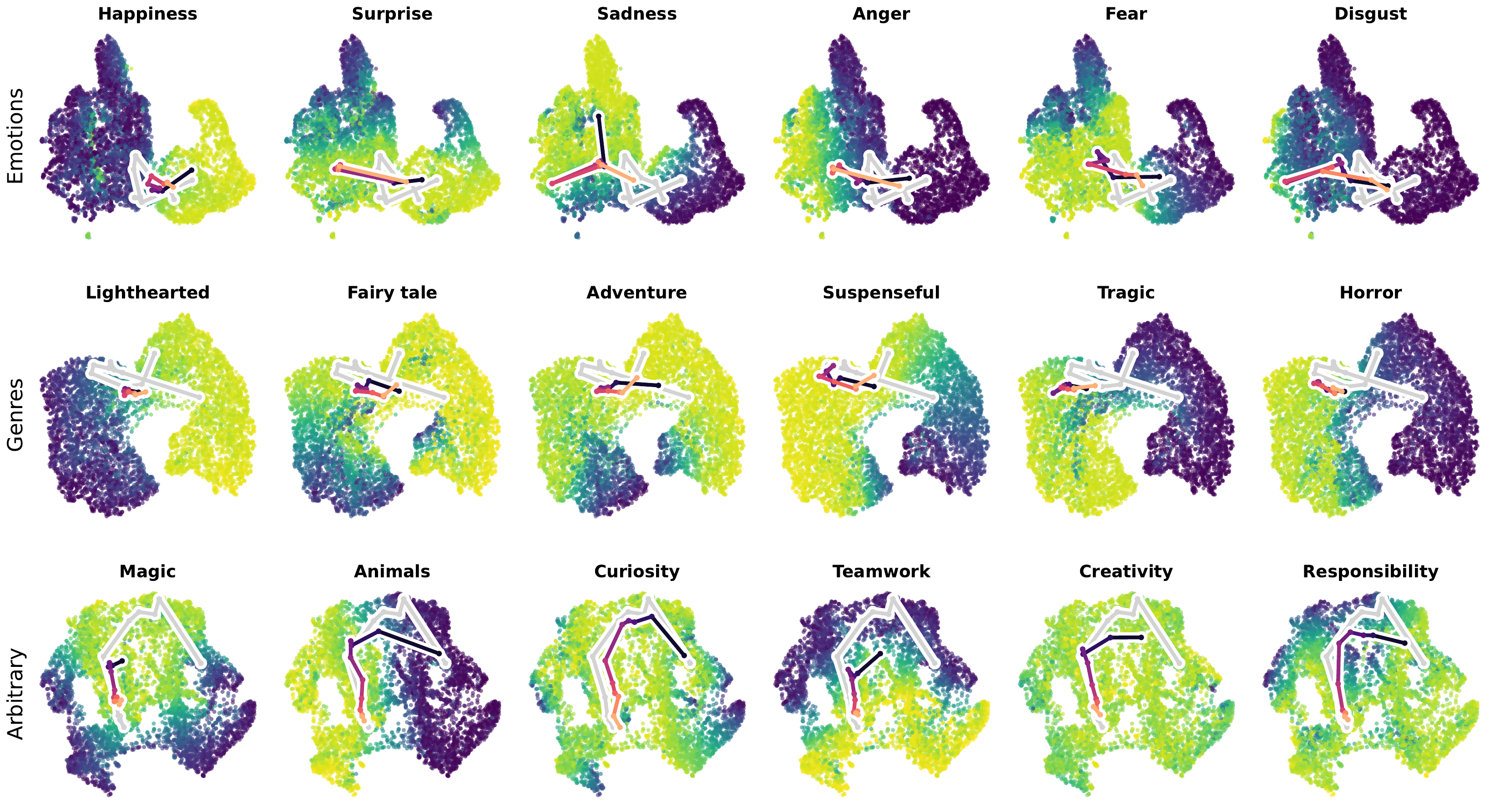}
    \caption{Steering effects for each individual concept $c$ with linear probe, projected into $\mathcal{M}_y$ for each domain}
    \label{fig:steer-all-beh} 
\end{figure}

\begin{figure}[t]
    \centering
    \includegraphics[width=\linewidth]{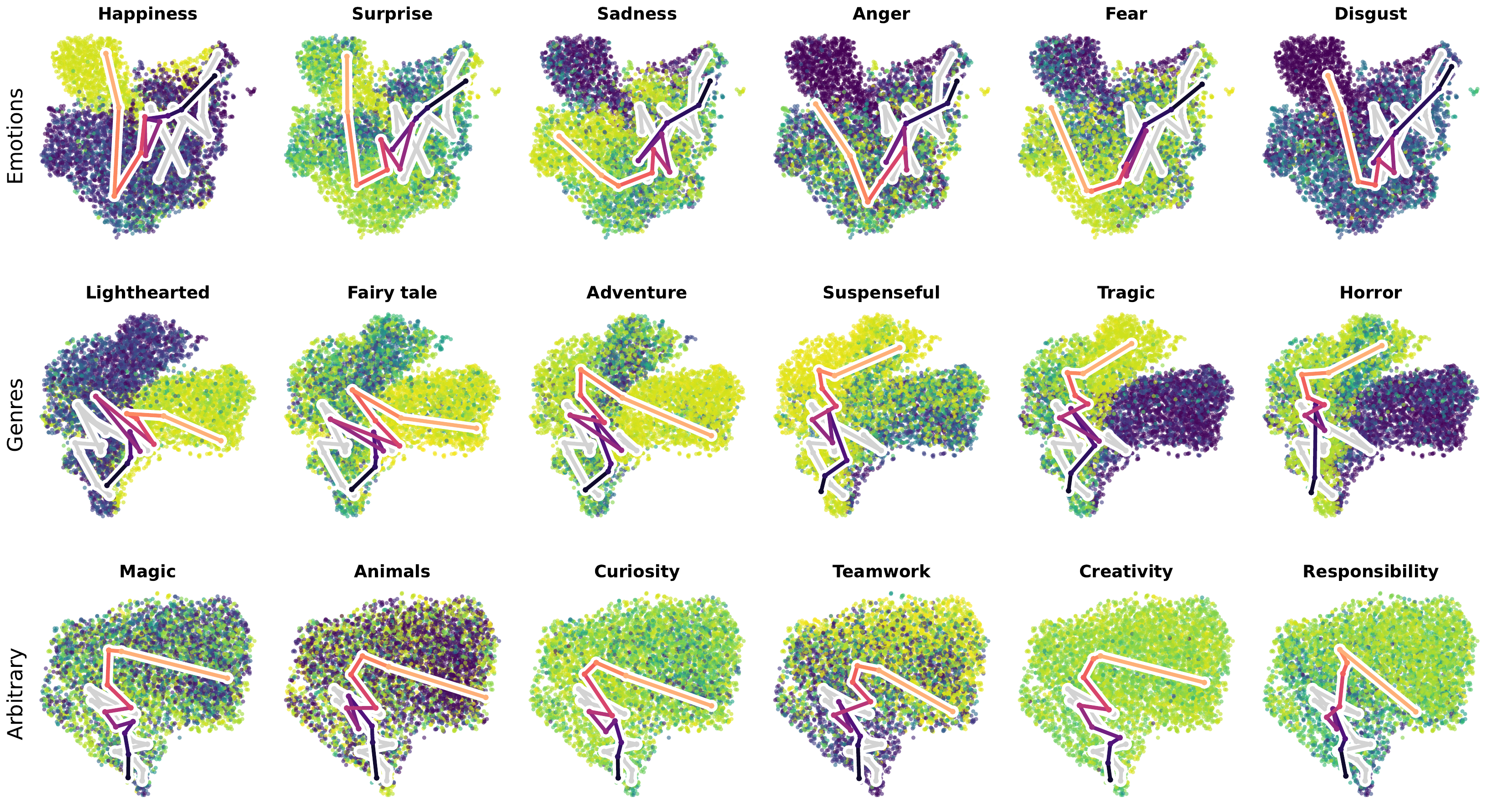}
    \caption{Steering effects for each individual concept $c$ with linear probe, projected into $\mathcal{M}_h$ for each domain}
    \label{fig:steer-all-rep} 
\end{figure}

Here, in Fig.~\ref{fig:steer-all-beh} and Fig.~\ref{fig:steer-all-rep} we present our main results in Figure~\ref{fig:steer-geom} for all three domains, and for each concept $c$ in each domain.

\newpage
\section{Main Analyses for Arbitrary Domain} ~\label{app:arbitrary-analysis}

\begin{figure}[t]
    \centering
    \includegraphics[width=\linewidth]{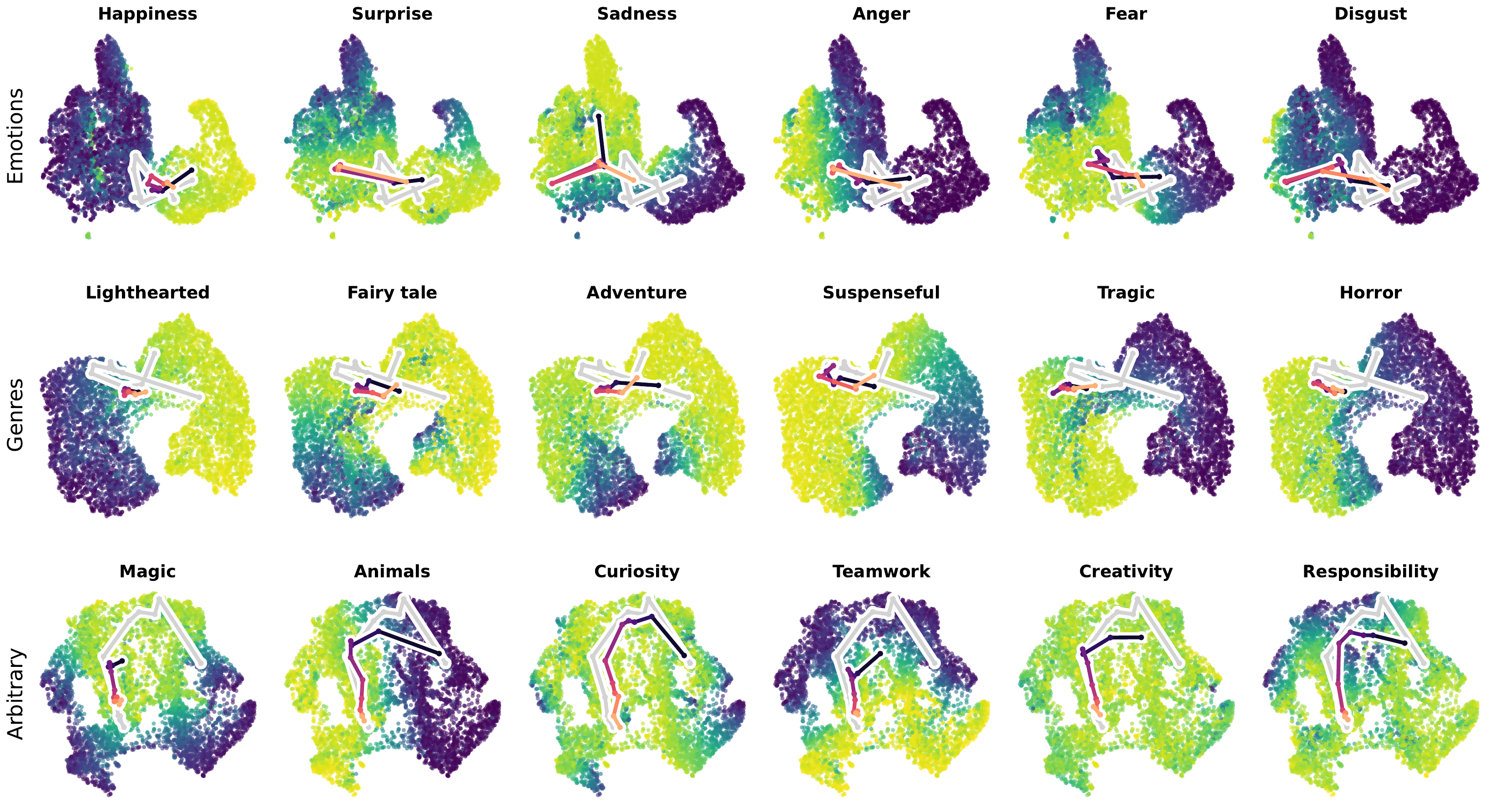}
    \caption{Distance matrix results at in Fig.~\ref{fig:distance-mats} for \textit{Arbitrary} Domain.}
    \label{fig:arb-dists} 
\end{figure}

\begin{figure}[t]
    \centering
    \includegraphics[width=\linewidth]{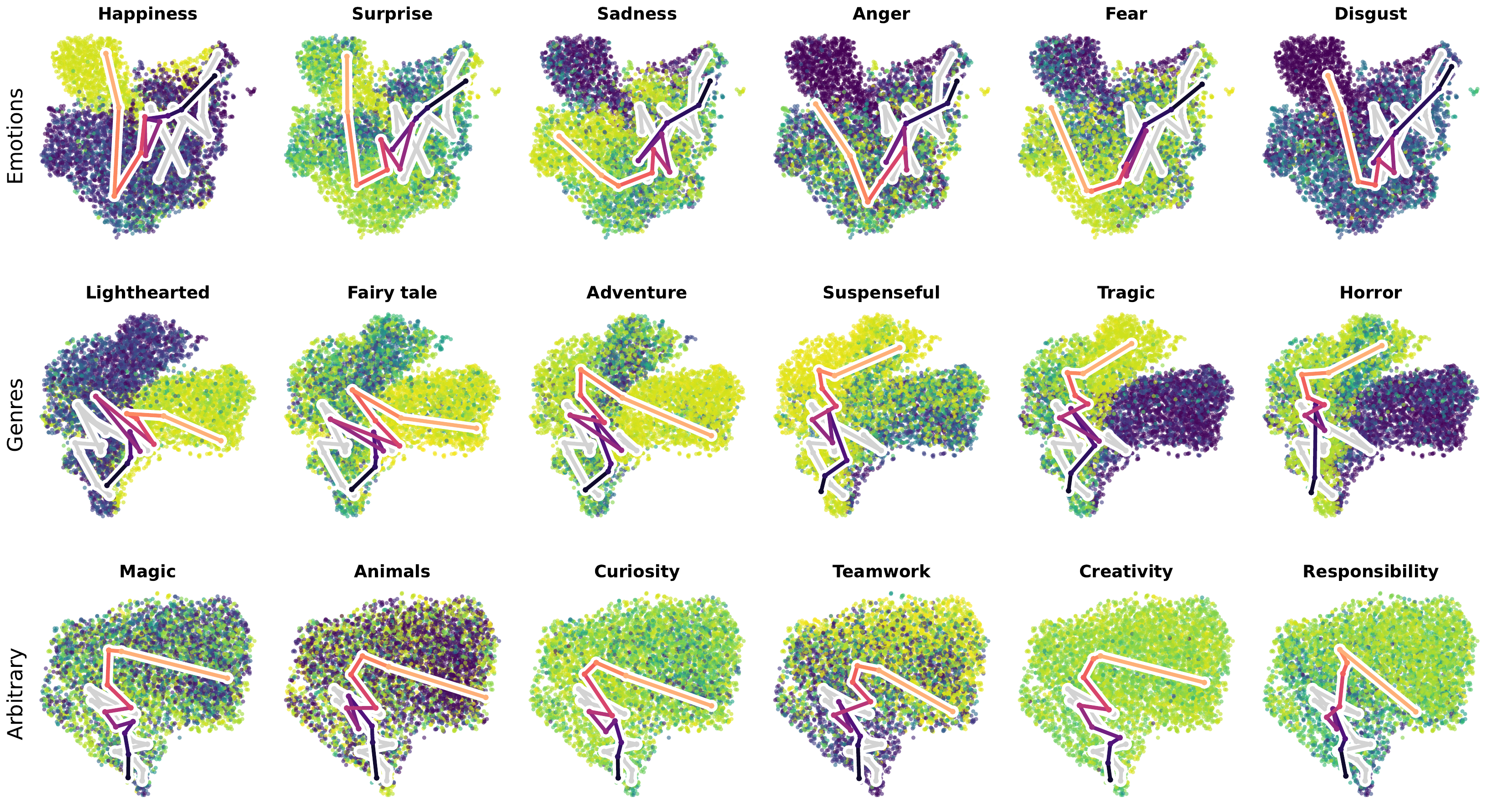}
    \caption{Steering entanglement and prediction from $\mathcal{M}_y$ for \textit{Arbitrary} domain, as in Fig.~\ref{fig:steering-entangle}.}
    \label{fig:arb-steer} 
\end{figure}

We show distance matrix analysis (Fig.~\ref{fig:arb-dists}) and steering entanglement (Fig.~\ref{fig:arb-steer}) for the \textit{Arbitrary} domain.

We find that, while $\mathcal{M}_z$ has clear structure in the distance matrix, this structure is less similar to $\mathcal{M}_y$ than is the case for our other two domains. We also find that structure in $\mathcal{M}_y$ is not predictive of steering entanglement in this domain, as it is for \textit{Emotions} and \textit{Genres}. 

Taken together, these results support our hypothesis that a relatively random set of concepts grouped together will not satisfy our expected criteria for being a \textit{domain}. That said, it is not surprising that we find structure in $\mathcal{M}_z$ and $\mathcal{M}_y$ given that there are likely various subtler correlations between the concepts in the \textit{Arbitrary} domain and in the max-activating examples. For example, in our story dataset, \textit{animals} and \textit{magic} are loosely correlated since many stories with animals also involve talking animals, and thus magic.

\newpage
\section{Selecting Optimal Layer for Probe} ~\label{app:probe-opt-layer}

\begin{figure}[t]
    \centering
    \includegraphics[width=\linewidth]{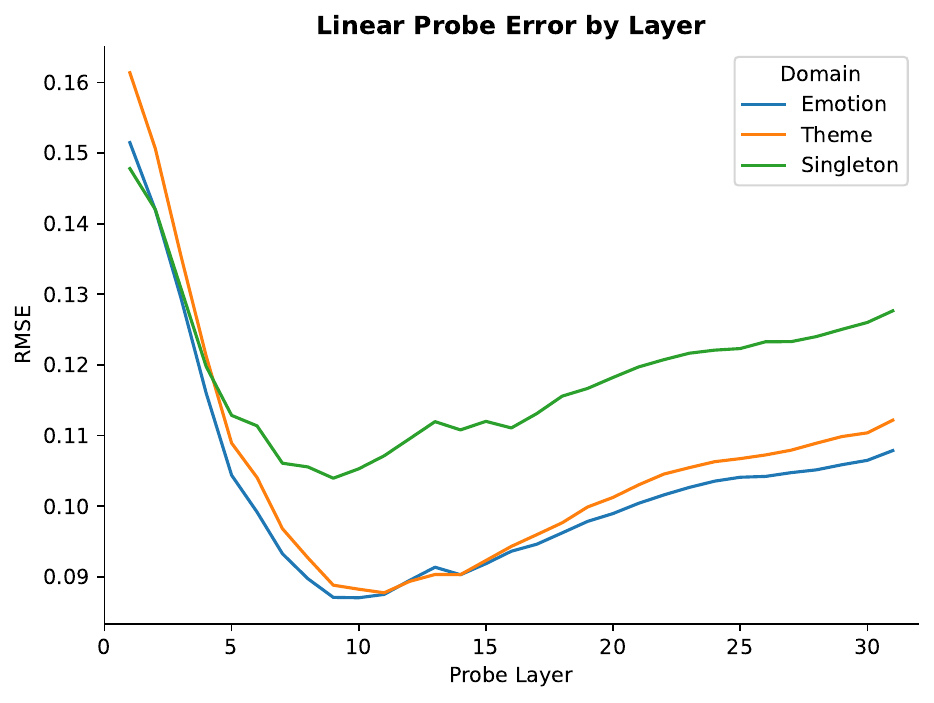}
    \caption{Linear accuracy for each layer, across each domain.}
    \label{fig:probe-acc-by-layer} 
\end{figure}

We use layer 9 in our steering experiments, which we select as the layer where linear probes for all domains are approximately optimal, as shown in Fig.~\ref{fig:probe-acc-by-layer}

\newpage
\section{Probe Steering Effects by Layer} ~\label{app:probe-effect-repair}

We empirically find that steering with linear probe weights at a single layer does not change model behavior. To better understand why this happened, we used our linear probes for a set of layers $\ell$ spaced throughout the model to measure whether steering effect actually occurs, but disappears at later layers due to model self-repair.

\begin{figure}[t]
    \centering
    \includegraphics[width=\linewidth]{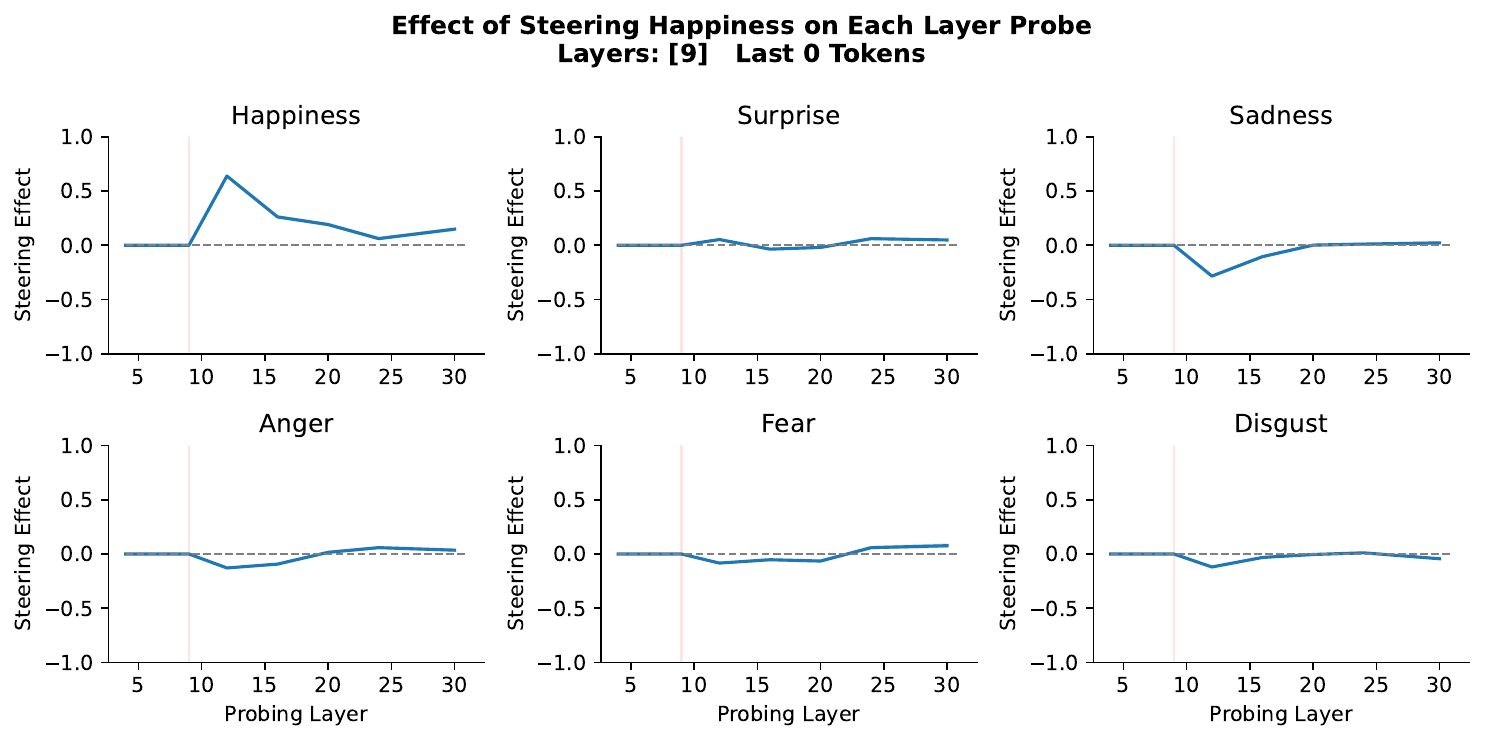}
    \caption{Using probe weights to steer at layer 8 only, we see that steering effect disappears after a few layers. Note: layer indexes in title are off by 1.}
    \label{fig:probe-repair-single} 
\end{figure}

\begin{figure}[t]
    \centering
    \includegraphics[width=\linewidth]{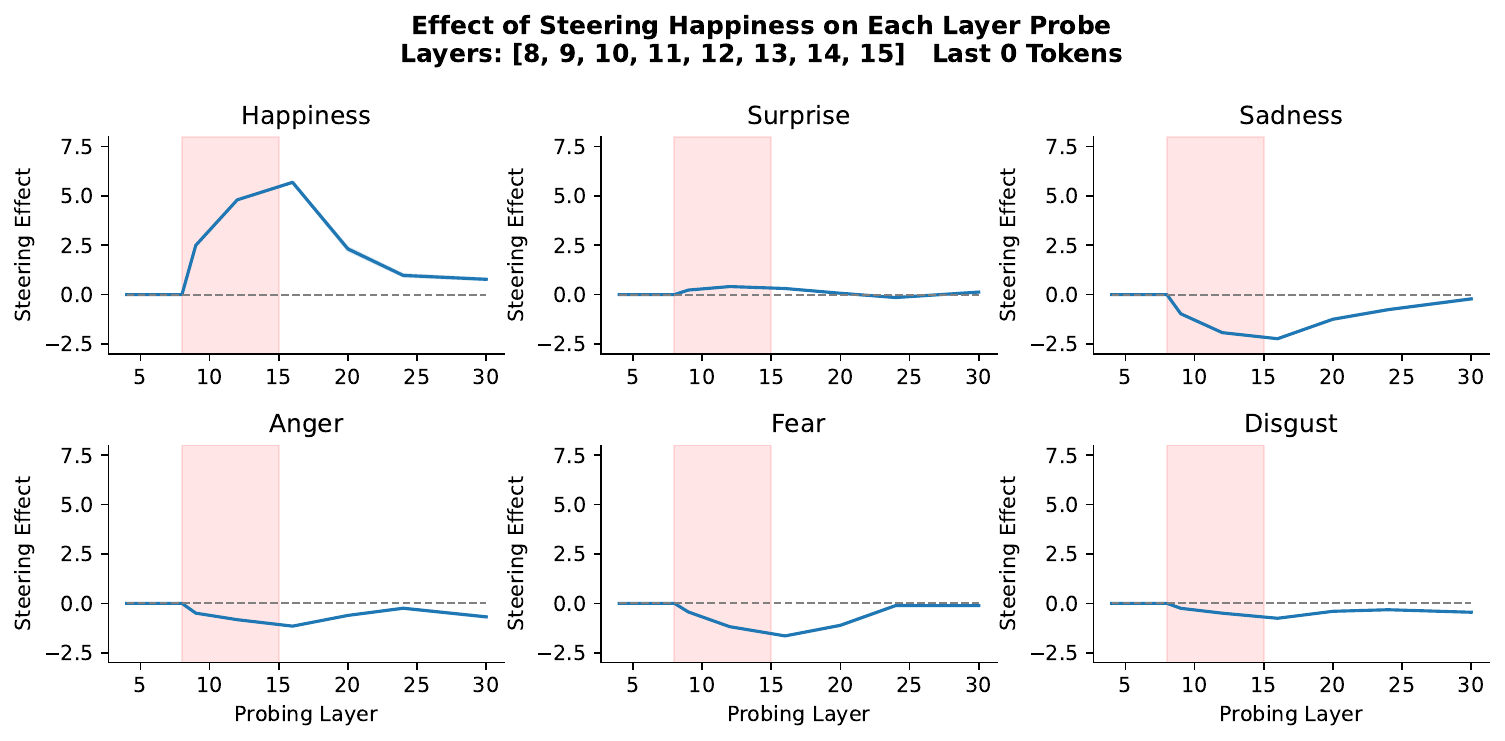}
    \caption{Using probe weights to steer at layers 7-14, we see that steering effect perseveres until the output layer. Note: layer indexes in title are off by 1.}
    \label{fig:probe-repair-multi} 
\end{figure}

We find that indeed, with one layer, steering effect occurs at the targeted layer but disappears quickly thereafter (Fig.~\ref{fig:probe-repair-single}). If we instead steer at a range of layers, the effect is able to persevere all the way to the model output (Fig.~\ref{fig:probe-repair-multi}).

\newpage
\section{Behavioral Data Distributions}  ~\label{app:raw-data}

\begin{figure}[h]
    \centering
    \includegraphics[width=\linewidth]{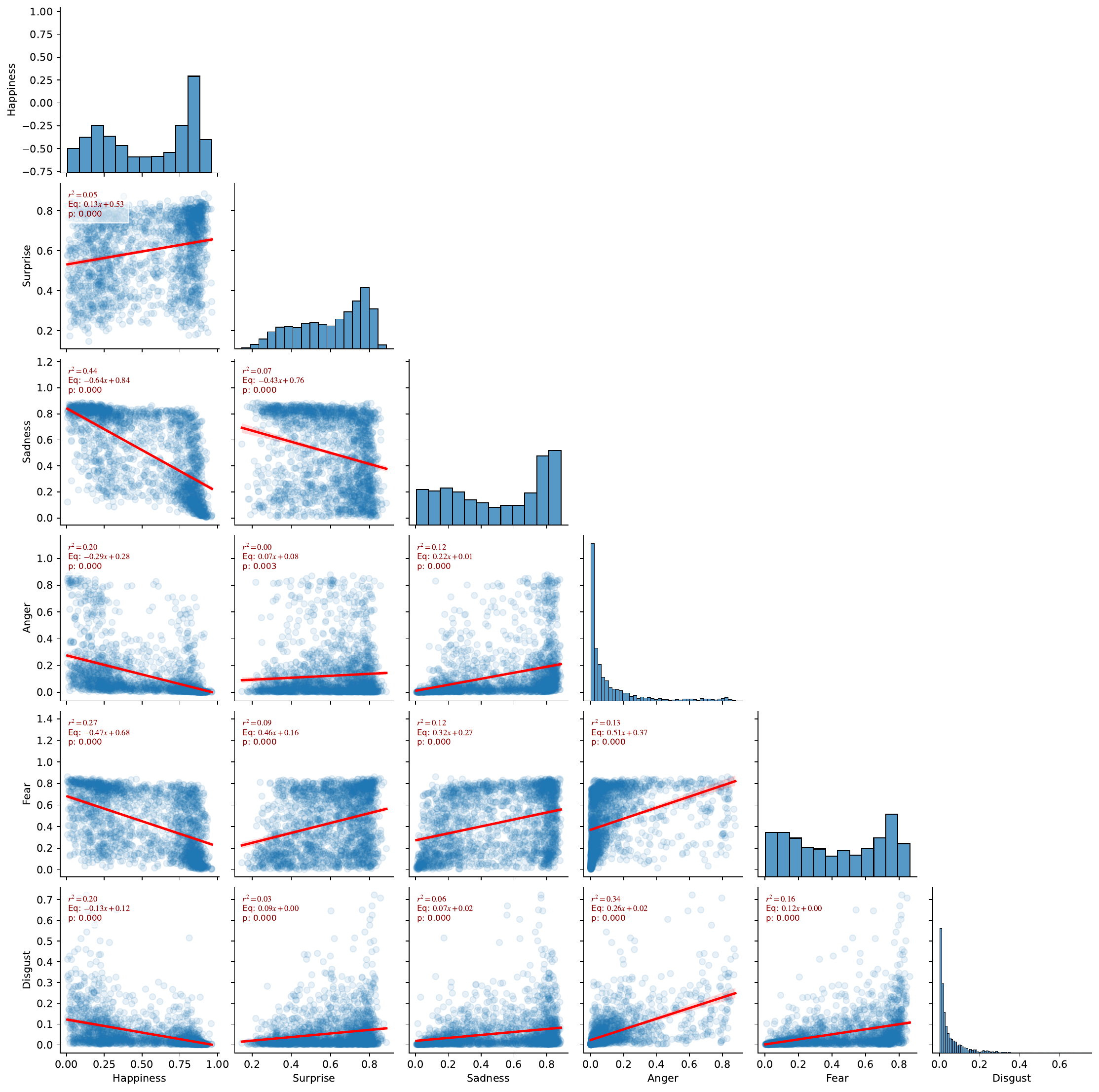}
    \caption{Correlations in $y_{t,c}$ for each pair of concepts in the \textit{Emotions} domain. A subset of 2000 random data points are shown here.}
    \label{fig:corrs-emo} 
\end{figure}

\begin{figure}[h]
    \centering
    \includegraphics[width=\linewidth]{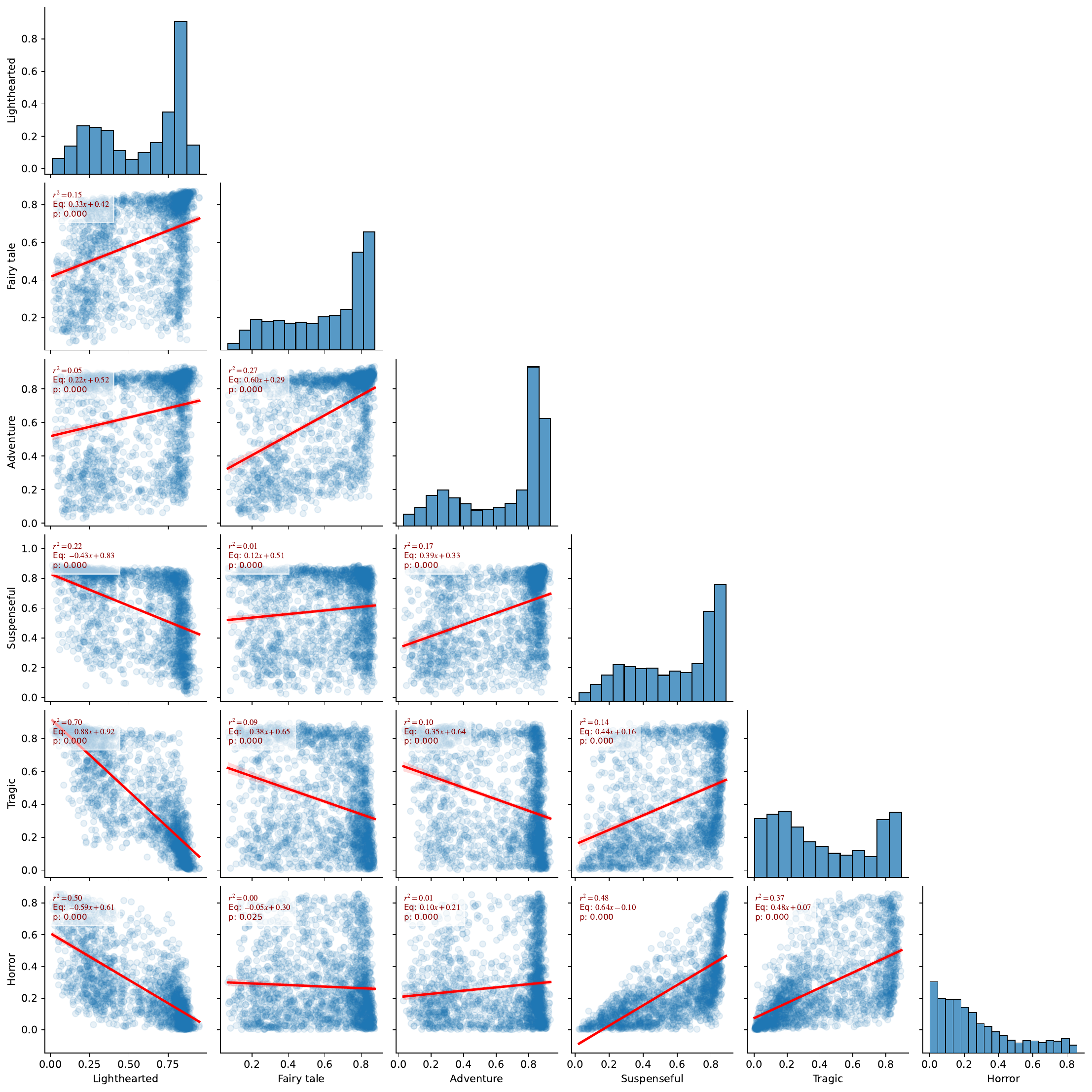}
    \caption{Correlations in $y_{t,c}$ for each pair of concepts in the \textit{Genres} domain. A subset of 2000 random data points are shown here.}
    \label{fig:corrs-genre} 
\end{figure}

\begin{figure}[h]
    \centering
    \includegraphics[width=\linewidth]{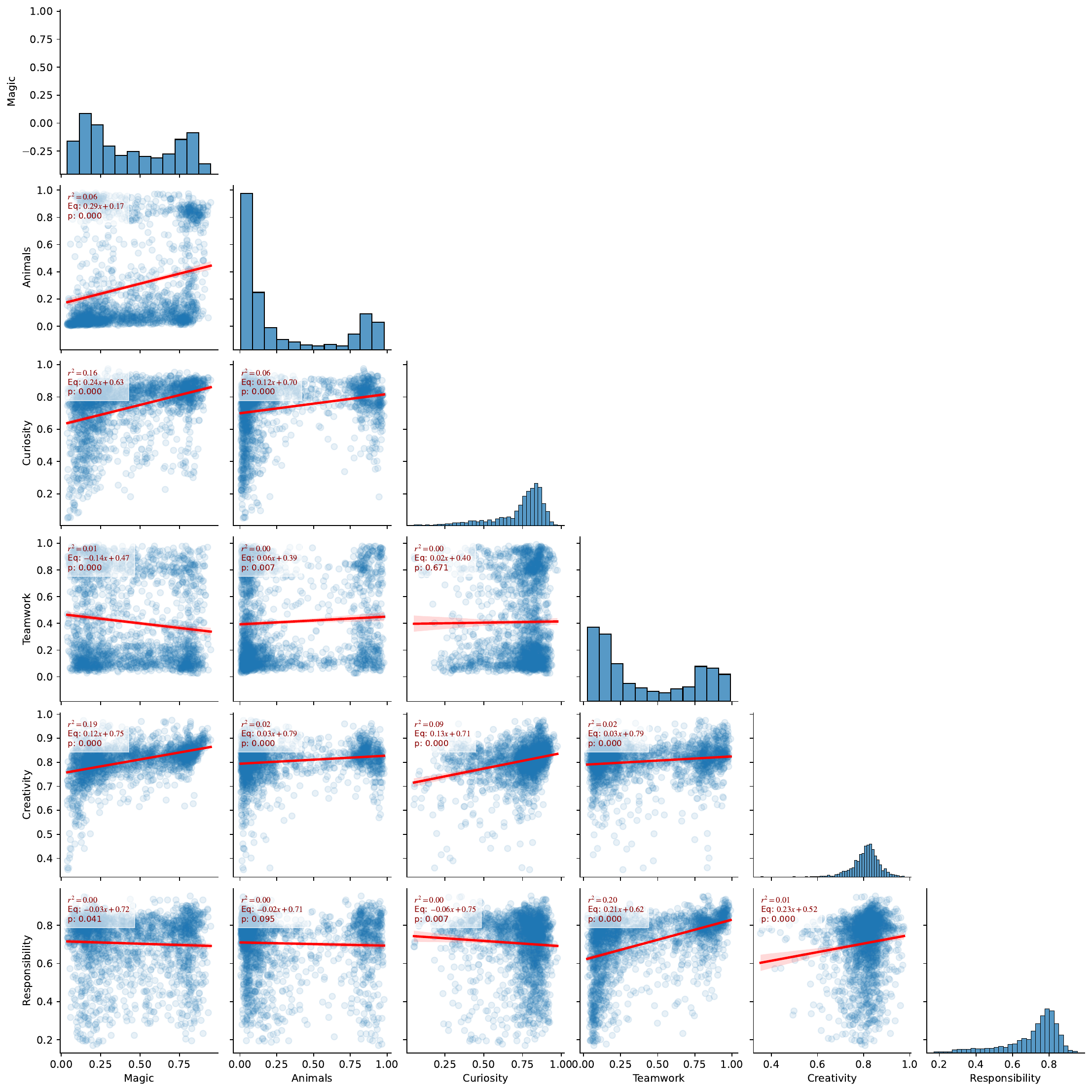}
    \caption{Correlations in $y_{t,c}$ for each pair of concepts in the \textit{Arbitrary} domain. A subset of 2000 random data points are shown here.}
    \label{fig:corrs-arb} 
\end{figure}

As a basic metric of how correlated our raw data $y_{t, c}$ is across concepts, in Figs.~\ref{fig:corrs-emo},~\ref{fig:corrs-genre},~\ref{fig:corrs-arb} we show correlations and histograms for each individual concept $c$ and each pair of concepts across all domains.

\newpage
\section{Calibrating Linear Probes}  ~\label{app:calibration}

\begin{figure}[h]
    \centering
    \includegraphics[width=.45\linewidth]{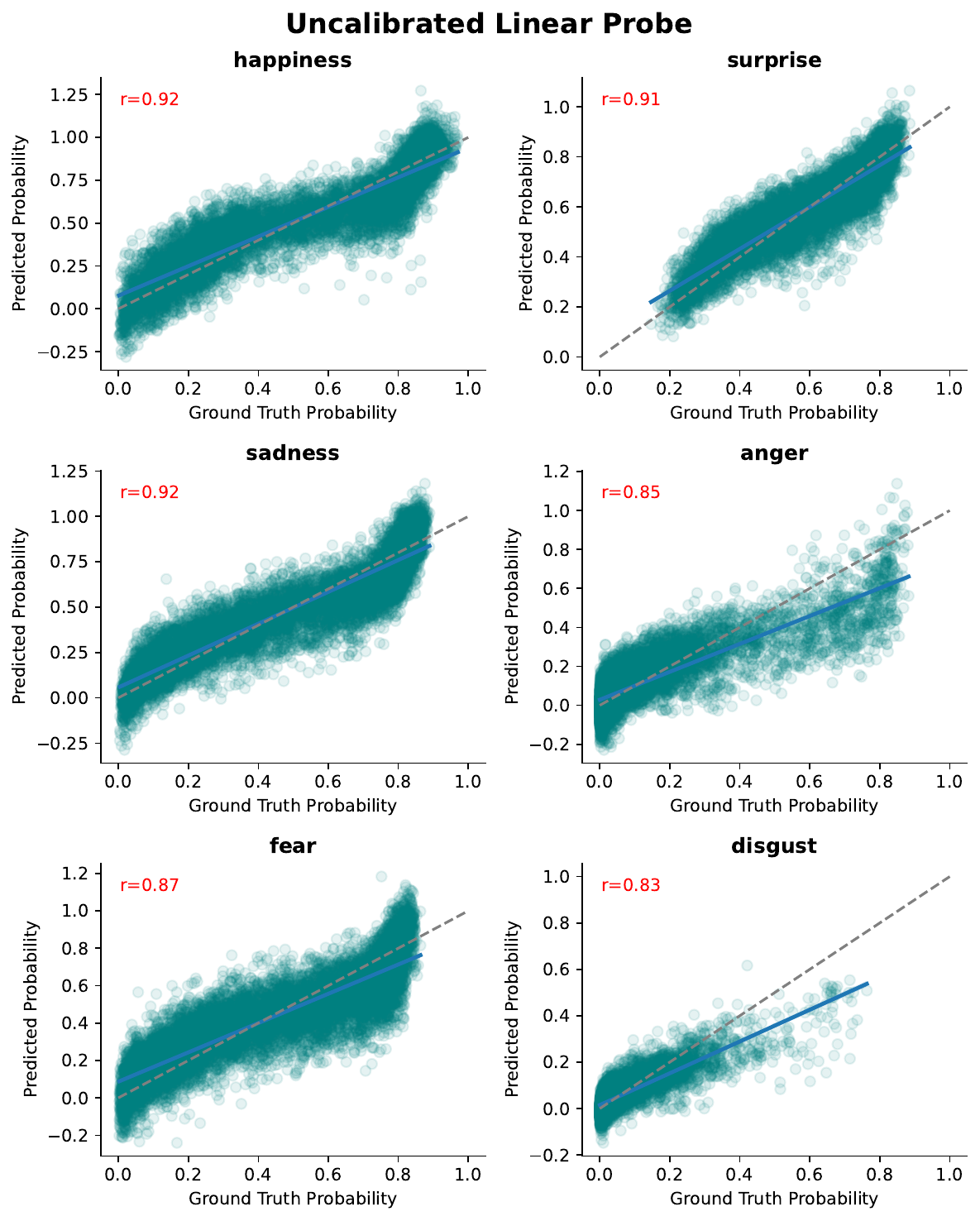}
    \quad
    \includegraphics[width=.45\linewidth]{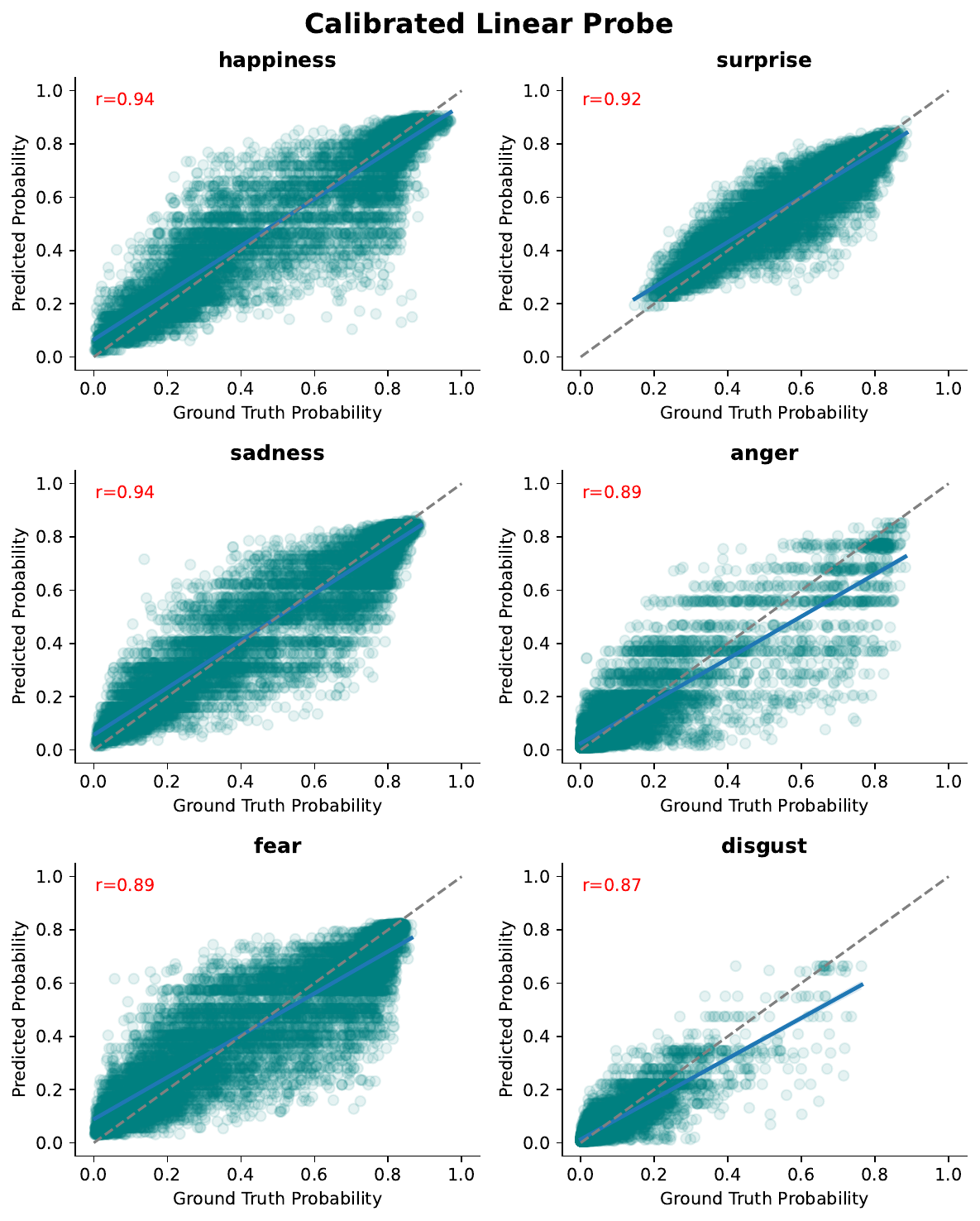}
    \caption{(Left) We find that probe predictions $\widehat{y}_{t,c}$  are initially miscalibrated with respect to ground truth  $y_{t,c}$. (Right) After applying isotonic regression, predictions are more calibrated and less systematically biased.}
    \label{fig:calibration} 
\end{figure}

We empirically observed that when plotting probe predictions $\widehat{y}_{t,c}$ against ground truth $y_{t,c}$, predictions were systematically biased, indicative of miscalibration. To rectify this, we applied isotonic regression, which largely corrects this bias (Fig.~\ref{fig:calibration}).

\newpage
\section{Further Analysis of Steering Effects}  ~\label{app:further-steering}

\begin{figure}[h]

    \centering
    \includegraphics[width=.65\linewidth]{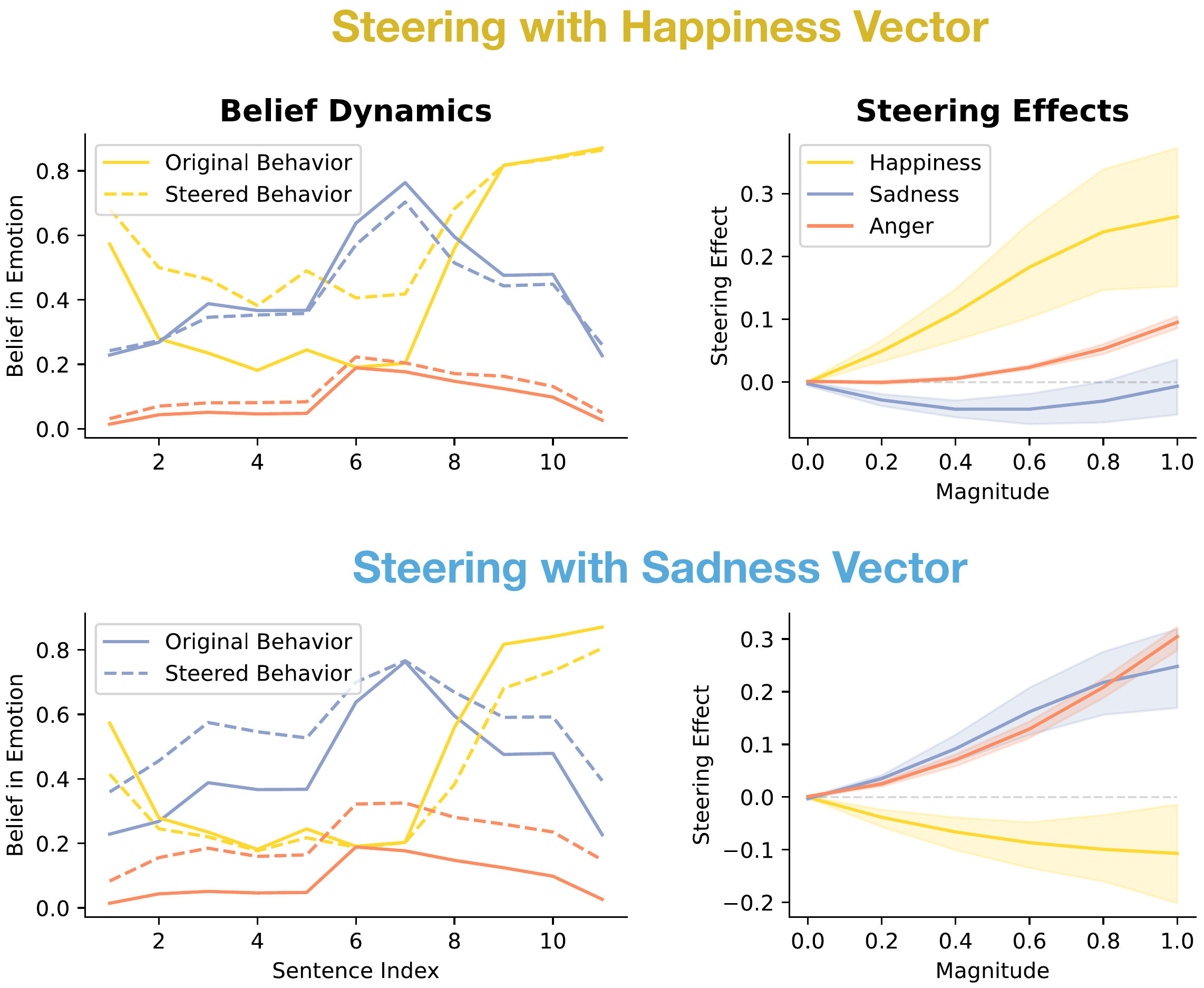}
    \caption{
        (Left) We observe that activation steering alters model belief dynamics, depending on which concept is targeted.
        (Right) Steering effect varies systematically as a function of steering vector magnitude.
        We also observe entanglement where steering effects unintended concepts, e.g. steering for \textit{sadness} impacts belief in \textit{anger}.
    }
    \label{fig:steering-dynamics}
\end{figure}

In Fig.~\ref{fig:steering-dynamics} we show the effects of steering for two concepts \textit{happiness} and \textit{sadness} on belief dynamics in $y_{t,c}$, as well as the effect of steering at varying magnitude $\alpha$. Results shown here are for the linear probe.

\newpage
\section{Belief Dynamics for Additional Stories}  ~\label{app:additional-stories}

\begin{figure}[h]
    \centering
    \includegraphics[width=\linewidth]{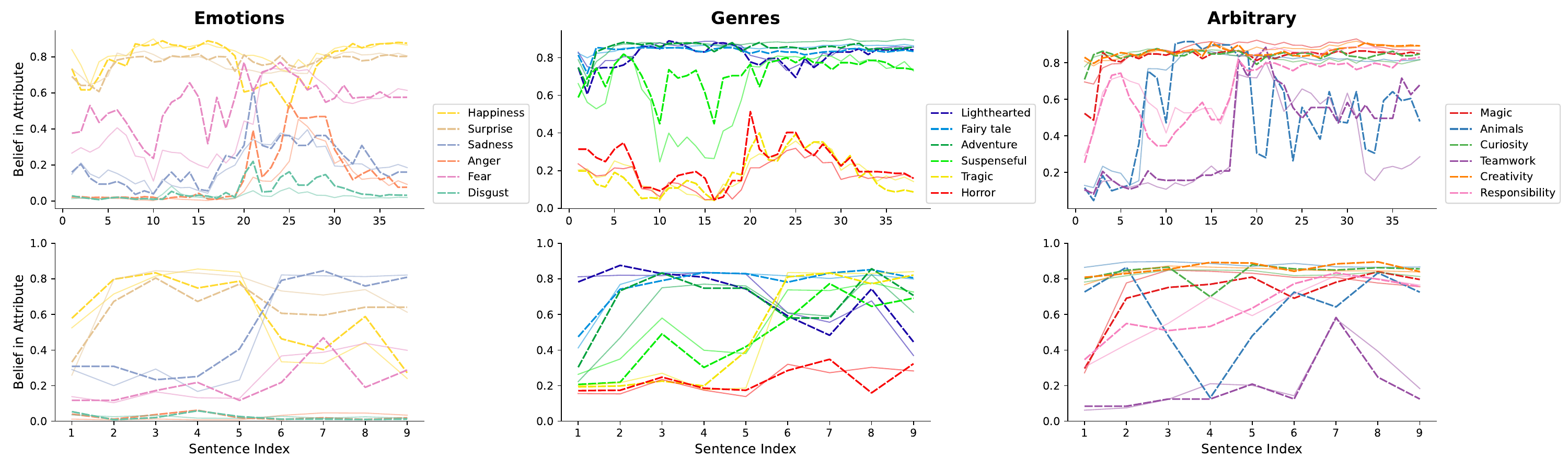}
    \vspace{-10pt}
    \caption{Belief Dynamics in $y_{t,c}$ for two additional stories.
    }
    \label{fig:two-other-stories} 
\end{figure}

The belief dynamics for each input text has its own nuanced story to tell. Here, in Fig.~\ref{fig:two-other-stories} we show two additional stories other than the one story considered in the main text. An interactive dashboard with belief dynamics and trajectories for all stories in our test set is available at: \url{https://conceptual-beliefs.streamlit.app}.

\newpage
\section{Sentence Index Encoded in Activation Manifold}  ~\label{app:sent-idx-mani}

\begin{figure}[h]
  \centering
  \includegraphics[width=0.5\linewidth]{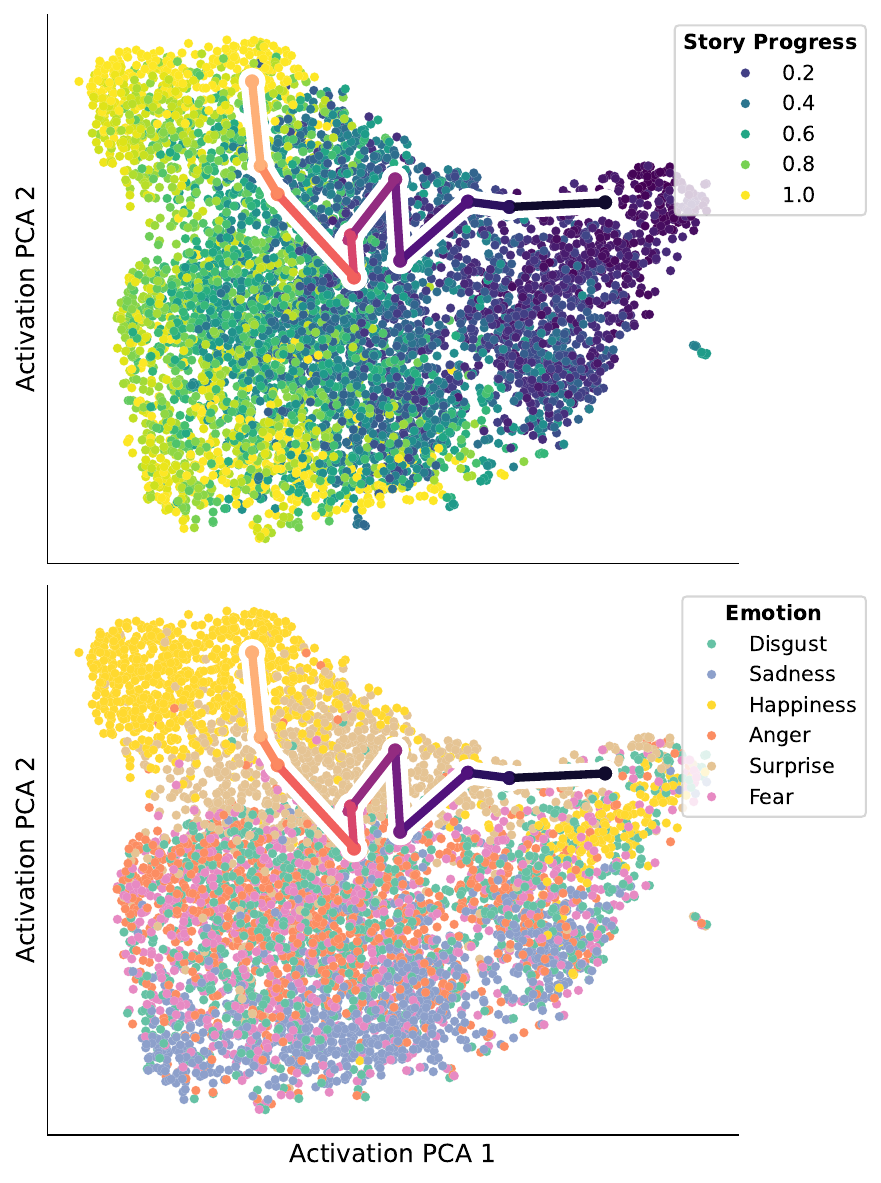}   
  \caption{PCA of activations for max activating examples, colored by the story progress, i.e., sentence index (Top) and by max-activating emotion label (Bottom)}
  \label{fig:embeds-simple}
\end{figure}

We find that $\mathcal{M}_z$ not only encodes the structure of the domain, as in Fig.~\ref{fig:steer-geom}, but it also encodes sentence index $t$, i.e., an estimate of the current progress within a story (Fig~\ref{fig:embeds-simple}).

\end{document}